\pdfoutput=1
\documentclass[letterpaper]{article}
\ifdefined\aaaianonymous
    \usepackage[submission]{aaai2026}
\else
    \usepackage{aaai2026}
    \nocopyright
\fi

\usepackage{times}
\usepackage{helvet}
\usepackage{courier}
\usepackage[hyphens]{url}
\usepackage{graphicx}
\usepackage{natbib}
\usepackage{caption}
\usepackage{amsmath,amssymb}
\usepackage{array}
\usepackage{booktabs,multirow}
\usepackage{algorithm}
\usepackage{algorithmic}

\newif\ifwithappendix
\withappendixtrue
\newcommand{\appsec}[1]{\ifwithappendix Appendix~\ref{#1}\else the supplementary material\fi}
\newcommand{\apptab}[2][]{\ifwithappendix Appendix Table~\ref{#2}#1\else the supplementary material\fi}

\newcommand{\seeapptab}[2][]{\ifwithappendix Appendix Table~\ref{#2}#1\else see the supplementary material\fi}

\newcommand{\tractus}{Tractus}
\newcommand{\tractusdec}{Tractus-decision}
\newcommand{\thdr}[1]{\begin{tabular}{@{}c@{}}#1\end{tabular}}

\title{Tractus: Fast Dual-Mode Decision Foundation Models\\with Temperature-Guided Deliberation}

\ifdefined\aaaianonymous
\author{Anonymous Submission}
\affiliations{}
\else
\author{
    Shaocong Xie,
    Zhiming Fang
}
\affiliations{
    Business School, University of Shanghai for Science and Technology, Shanghai, China\\
    xieshaocongethan@gmail.com, zhmfang2015@163.com
}
\fi

\begin{document}

\maketitle

\begin{abstract}
Autoregressive decoding incurs sequential latency and token overhead in time-critical decisions, and free-form text lacks calibrated action probabilities. Prior approaches face distinct tradeoffs: single-pass scorers run at fixed depth, disjoint cascades drop KV prefixes across models, and unconditional chain-of-thought decodes traces on routine queries. We introduce \textbf{Tractus}, a 0.8B--9B dual-mode family on Qwen3.5 sharing one checkpoint per scale: \tractusdec\ scores candidates in one prefill pass without decoding via a 99.9\% smaller readout, while \tractus\ continues over the same KV prefix into reasoning when needed. Mode selection uses \textbf{Thermostasis}, an analytic mechanism with five fixed constants and zero fitted temperatures combining bounded logit softcapping, shift-invariant centering, an analytic temperature schedule, joint confidence-margin routing, a safety ratchet, and two optional Decision Index~0.3 priors. Within our family under greedy decoding, adaptive deliberation emerges at 4B alongside Stage-3 RLVR and 32-layer depth: on 8K JevBench-231, Tractus raises 4B accuracy from 76.19\% to 88.31\% with fewer tokens than unconstrained reasoning and reaches 90.91\% at 9B, trailing unconstrained reasoning by 3.46\,pp while exiting in a single pass on most queries and cutting tokens by 38.1\%. Out of distribution at 4B--9B, unit-temperature softcapping yields the lowest tested single-pass calibration error, whereas the two-dimensional thermostat raises OOD error and serves mainly for triage while reducing 0.8B Stage-1 in-distribution ECE by 58.5\%. \tractusdec\ leads Decision Index~0.3 Balanced Skill across all four size tiers using two 9B priors while trailing on Balanced Raw. On the exploratory 831-task suite comparing 3-prompt Tractus against 1-prompt baselines, Tractus-9B resolves over half of queries without decoding and substantially lowers median and tail latency relative to Qwen3.5-9B Thinking with zero extraction failures via logit recovery, while Gap-Only routing reaches 92.2\% on JevBench-231.
\end{abstract}

\section{Introduction}
\label{sec:intro}

Many real-world decision scenarios require discrete action selection under tight latency budgets, where routine queries admit direct resolution while a harder subset involves conflicting rules or multi-step dependencies \cite{liu2026decision1,dhankhar2026drex}. Although large language models (LLMs) are increasingly deployed in these domains, autoregressive decoding is typically memory bandwidth bound over growing KV caches \cite{leviathan2023speculative,kwon2023vllm}, and free-form text lacks action constraints and calibrated probabilities \cite{guo2017calibration,duan2026jeval}. Such deployments need fast calibrated probabilities without decoding on routine inputs alongside multi-step deduction when rules conflict \cite{wei2022cot}.

\begin{figure}[t]
    \centering
    \includegraphics[width=\columnwidth]{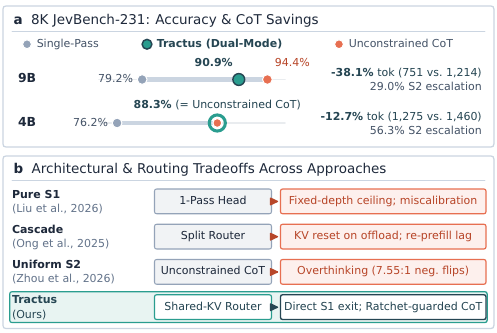}
    \caption{Accuracy--compute tradeoff on 8K JevBench-231 and architecture comparison. Top: vs.\ unconstrained CoT, Tractus cuts tokens by $12.7\%$ at 4B ($56.3\%$ esc.; matched $88.31\%$) and $38.1\%$ at 9B ($29.0\%$ esc.; $90.91\%$ vs.\ $94.37\%$, $-3.46$\,pp). Bottom: single-pass, disjoint cascade, unconstrained CoT, and Tractus.}
    \label{fig:teaser}
\end{figure}

Three active paradigms address only part of this gap (Figure~\ref{fig:teaser}). Single-pass scorers avoid decoding \cite{liu2026decision1,dhankhar2026drex}, but fixed-depth transformers face constant-depth expressivity ceilings \cite{merrill2023parallelism}, overstate top-choice probabilities \cite{duan2026jeval}, and miscalibrate under shift \cite{ovadia2019trust}. Disjoint routers and cascades offload uncertain queries \cite{ong2024routellm,wu2026reflex}, dropping KV state while BFCL tool-relevance detection stays at 52.0\% despite 98.4\% function selection \cite{wu2026reflex}, and weight interpolation blends two autoregressive checkpoints without single-pass exits \cite{yang2026dami}. Unconditional reasoning LLMs often decode long traces on routine queries \cite{guo2025deepseekr1}, and can overthink with a 7.55:1 negative-to-positive flip ratio on competition math under budget forcing \cite{zhou2026overthinking}.

One structural factor underlies these tradeoffs: separating fast scoring, deliberation, and calibration across disjoint models or post-hoc stages. Inspired by dual-process cognition \cite{kahneman2011thinking}, our goal is one shared-weight model that scores routine queries in a single prefill pass (System~1), escalates uncertain cases over the live KV prefix into CoT (System~2), and calibrates without fitted temperatures.

A single-checkpoint architecture creates coupled training and inference difficulties. Jointly tuning one backbone for single-pass classification and multi-step CoT introduces \emph{representation interference}, as autoregressive reasoning updates can degrade mid-layer single-pass circuits. Even when those circuits remain intact, skewed binary decision tasks induce \emph{binary prior bias}, imprinting prior logit offsets that push standard objectives toward majority shortcuts \cite{zhao2021calibrate}. At inference time, unbounded raw logits further suffer from \emph{OOD overconfidence}, while validation-fitted scalar temperatures often degrade under distribution shift \cite{ovadia2019trust}, leaving static routing thresholds prone to either under-escalating compositional queries or triggering overthinking on routine ones.

We introduce \textbf{Tractus} (0.8B--9B on Qwen3.5) and \textbf{Thermostasis}, an analytic mechanism with zero fitted temperatures, five fixed constants, and two optional Decision Index~0.3 (DI~0.3) priors. Our main contributions are:
\begin{itemize}
    \item \textbf{Shared-Prefix Dual-Mode Architecture (Section~\ref{sec:arch}):} A 0.8B--9B Qwen3.5 family (256K capacity; 64K--256K untested) uniting a 99.9\% smaller candidate-restricted readout (\tractusdec) and live-KV prefix-reusing CoT (\tractus) in one checkpoint per scale without state discard.
    \item \textbf{Thermostasis Calibration and Triage (Sections~\ref{sec:thermostasis}, \ref{sec:ablations}):} Zero fitted temperatures pairing bounded logit softcapping, shift-invariant centering, Joint or \textbf{Gap-Only (GO)} routing, and a safety ratchet \cite{gemma2024gemma2}; ablations separate softcapping OOD calibration from 2D thermostat triage, while Thermostasis cuts 0.8B Stage-1 ID ECE by 58.5\%.
    \item \textbf{Anti-Interference and Debiased Post-Training (Sections~\ref{sec:training}--\ref{sec:benchmark}):} Stage-1 distillation, Stage-2 tiered-LR single-pass replay protecting mid-layer circuits, and Stage-3 interactive and Antithetic-Pair RLVR (ATP-RLVR) on positive--negative twins (4B/9B only) curbing prior bias. Within our family under greedy decoding, adaptive deliberation emerges at 4B alongside Stage-3 RLVR and 32-layer depth, matching unconstrained CoT on 8K JevBench-231 at 4B and cutting tokens by 38.1\% at 9B with a 3.46\,pp gap; \tractusdec\ leads DI~0.3 Balanced Skill across all four tiers (two 9B priors; trailing Balanced Raw), while 3-prompt Tractus-9B resolves over half of the 831-task exploratory suite without decoding at far lower probe latency and zero logit-recovery failures than 1-prompt Qwen3.5-9B Thinking.
\end{itemize}

\section{Related Work}
\label{sec:related}

\paragraph{Decision Foundation Models and Dual-System Routing.}
Single-pass models---including Jev, vLLM-SR Decision, and Drex---score options in one pass \cite{duan2026jeval,liu2026decision1,dhankhar2026drex}, though fixed depth limits parallel expressivity \cite{merrill2023parallelism}. Disjoint routers and cascades drop KV state \cite{chen2023frugalgpt,ong2024routellm,saha2024system1x,wu2026reflex}, whereas weight interpolation, symbolic provers, and early-exit or speculative decoding do not unify single-pass scoring and adaptive CoT \cite{yang2026dami,hu2025ablrefl,banino2021pondernet,schuster2022calm,leviathan2023speculative}. Closest to Tractus, StartLux-Decision splits Instant (bidirectional) and AutoThink (causal) checkpoints with single-threshold routing, 0.25/0.75 blending, and $\text{RLCD}^2$ \cite{zhuang2026startlux}. Tractus unifies both in one hybrid Gated DeltaNet checkpoint \cite{yang2025gateddeltanet,yang2025qwen3}, mitigating interference via tiered-LR replay, prior bias via ATP-RLVR, and OOD overconfidence/overthinking via softcapping, margin routing, and a ratchet.

\paragraph{Adaptive Test-Time Scaling and Overthinking.}
CoT, test-time scaling, and RLVR aid deduction \cite{wei2022cot,wang2023selfconsistency,snell2024scaling,lambert2024tulu3,guo2025deepseekr1}, but uniform CoT can hurt simple queries \cite{alomrani2025budget}, and it can flip correct answers \cite{zhou2026overthinking,ghosal2025mirage}; brevity methods shorten traces during decoding \cite{hassid2025shorter,yang2025tops}, and sub-problem routing still decodes multi-token steps \cite{xiao2026scale}, whereas Thermostasis can skip decoding altogether and revert low-confidence rollouts.

\paragraph{Logit Calibration and Decision Biases.}
Choice scoring often exhibits label bias \cite{zhao2021calibrate}, order sensitivity \cite{zheng2024pride,hanna2026decoupling}, top-choice probability overstatement \cite{duan2026jeval}, and shift miscalibration \cite{ovadia2019trust}. Masking and permutation methods mitigate order flips \cite{mcilroy2024sbp,zheng2024pride,jiao2026ember,fatemi2026gps,zheng2026pagrpo}, whereas static scaling rescales margins uniformly \cite{guo2017calibration}; default single-pass Thermostasis ($\mathbf{z}_0{=}\mathbf{0}$) leaves cyclic flips invariant and pairs bounded softcapping with an analytic triage temperature \cite{gemma2024gemma2}.

\section{Methodology}
\label{sec:method}

\subsection{Problem Formulation and Unified Architecture}
\label{sec:arch}

Tractus accepts text, structured records, or visual inputs with a dynamic menu of $2 \le K \le 255$ candidates, assigning each option a reserved single-token slot: ASCII uppercase letters for $K \le 26$ and non-acronym tokens above 26. At prefill, the backbone projects its final hidden state onto only those $K$ candidate rows to obtain raw System~1 logits $\mathbf{z}_{\mathrm{raw}}$ in one pass. When one option clearly leads, Thermostasis exits without decoding as \tractusdec; otherwise it can continue from the same prefix into a reasoning trace and System~2 answer as \tractus, allocating extra compute for deliberation without changing weights.

\begin{figure*}[t]
    \centering
    \includegraphics[width=\textwidth]{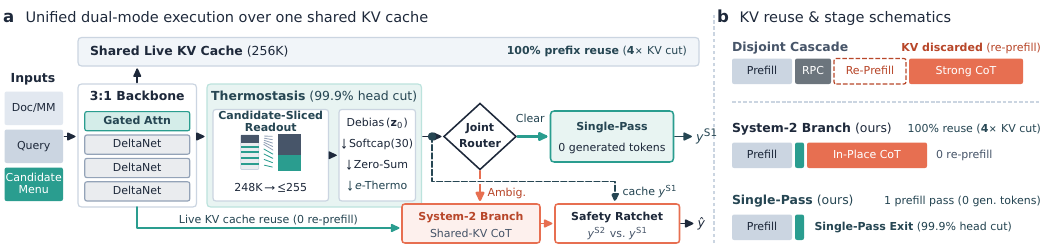}
    \caption{Unified dual-mode architecture of Tractus. (a)~A single 3:1 Gated DeltaNet and Gated Attention prefill feeds a $K$-candidate Thermostasis readout ($2 \le K \le 255$) for either a single-pass exit or adaptive System~2 deliberation with a safety ratchet. (b)~Unlike disjoint routers and cascades that route across separate models without prefix KV reuse \cite{ong2024routellm,wu2026reflex}, Tractus shares one checkpoint and prompt prefix across both modes.}
    \label{fig:architecture}
\end{figure*}

\begin{figure*}[t]
    \centering
    \includegraphics[width=0.96\textwidth]{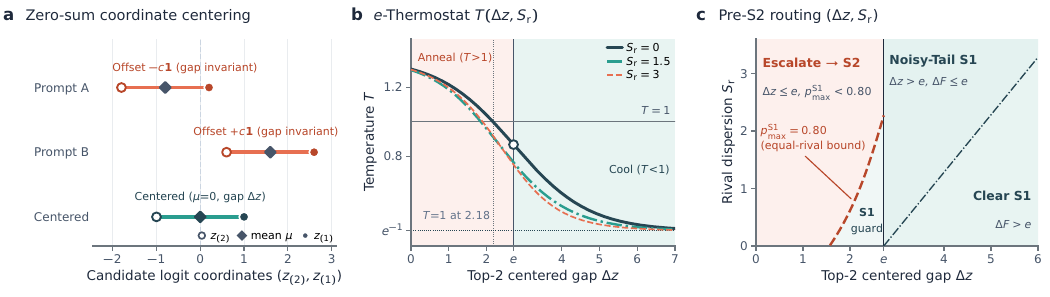}
    \caption{Geometry of Thermostasis calibration and routing (Section~\ref{sec:thermostasis}). (a)~Zero-sum centering via Eq.~\eqref{eq:centering} maps bounded logits $\bar{\mathbf{z}}$ to zero-mean canonical coordinates while leaving $\Delta z$ and softmax probabilities invariant. (b)~Analytic temperature $T(\Delta z, S_{\mathrm{r}})$ inflects at $\Delta z = e$ when rival dispersion $S_{\mathrm{r}}=0$, warming ties and cooling decisive leads. (c)~Diagnostic three-zone plane preserving Noisy-Tail System~1 exits; the dashed equal-rival boundary $\Delta z/T - S_{\mathrm{r}} = \ln 4$ is exact for $p_{\max}^{\mathrm{S1}}=0.8$ when all $K{-}1$ rivals tie.}
    \label{fig:geometry}
\end{figure*}

Tractus initializes from Qwen3.5-Base\footnote{\raggedright\url{https://qwen.ai/blog?id=qwen3.5}} at 0.82B, 1.88B, 4.21B, and 9.24B parameters (0.8B--9B). Each model interleaves three Gated DeltaNet layers per Gated Attention layer \cite{yang2025gateddeltanet}, with a 248,320-token vocabulary and an inherited 256K context capacity of 262,144 tokens. The two smaller models use 24 layers, and 4B and 9B use 32 layers. Inherited from Qwen3.5's hybrid layout, running full attention with 4 key-value heads of dimension 256 in only one layer in four reduces 32-layer BF16 attention KV-cache memory fourfold at 256K context, from $34.36$\,GB to $8.59$\,GB (with Gated DeltaNet's fixed-size recurrent state held separately); see Figure~\ref{fig:architecture}a. Images and video enter through the Qwen3.5 vision encoder; native audio waveforms are outside our current scope.

Prefill readout over only the $2 \le K \le 255$ candidate rows skips the full 248,320-way vocabulary projection, cutting final-token readout projection compute by ${\approx}99.9\%$ while reusing the same output matrix across both regimes. Sequential single-stream inference retains $100\%$ of the prefill KV prefix without re-prefilling (not separately wall-clock ablated here); high-throughput bulk evaluations instead decouple single-stream Transformers scoring from batched vLLM System~2 rollouts without cross-engine KV reuse; see Figure~\ref{fig:architecture}b and \appsec{sec:app_repro}.

\subsection{Thermostasis: Analytic Calibration and Adaptive Routing}
\label{sec:thermostasis}

Raw logits $\mathbf{z}_{\mathrm{raw}}$ over $K$ candidates can conflate evidence with large magnitudes and class priors. Thermostasis uses zero validation-fitted temperature scalars and five fixed constants ($C{=}30, e^{-1}, \tau_z{=}e, \tau_p{=}0.8, \tau_r{=}0.4$), plus an optional static binary prior $\mathbf{z}_0$ restricted to two 9B Decision Index~0.3 tasks ($\mathbf{z}_0{=}\mathbf{0}$ across 0.8B--4B, other DI~0.3 catalogs, primary benchmarks, and OOD splits). First, optional debiasing and Gemma~2 $\tanh$ softcapping yield $\bar{\mathbf{z}} = C \tanh((\mathbf{z}_{\mathrm{raw}} - \mathbf{z}_0)/C)$ \cite{gemma2024gemma2}, with dev-estimated $\mathbf{z}_0 = (+\Delta z_0/2,-\Delta z_0/2)$ at $\Delta z_0 = +3.2$\,nats on RAGTruth, $-0.75$\,nats on PhishNChips, and $\mathbf{z}_0{=}\mathbf{0}$ otherwise (\appsec{sec:app_repro}). Zero-sum centering over $j \in \{1,\dots,K\}$ gives $\tilde{\mathbf{z}}$ (Figure~\ref{fig:geometry}a):
\begin{equation}
    \tilde{z}_j = \bar{z}_j - \frac{1}{K}\sum_{k=1}^K \bar{z}_k.
    \label{eq:centering}
\end{equation}
Because $\Delta z$, $S_{\mathrm{r}}$, $T$, $\mathrm{softmax}$, $\arg\max$, and routing depend only on pairwise differences, Eq.~\eqref{eq:centering} is a shift-invariant coordinate canonicalization for geometry; it neither alters softmax nor removes raw shifts once $\tanh$ saturates, while softcapping at $C{=}30$ accounts for the observed $T{=}1$ OOD calibration gain.

Second, sorting $\tilde{\mathbf{z}}$ descending gives top-two gap $\Delta z = \tilde{z}_{(1)} - \tilde{z}_{(2)}$ and unit-temperature rival dispersion $S_{\mathrm{r}} = \log \sum_{j=2}^K \exp(\tilde{z}_{(j)} - \tilde{z}_{(2)})$ ($0$ for $K{=}2$). Rather than fitting validation temperatures, the analytic $e$-thermostat anchors at $\Delta z = e$ ($\exp(e) \approx 15.15:1$ binary odds), lower asymptote $e^{-1}$, and $T{=}1$ near $\Delta z \approx 2.18$ at $S_{\mathrm{r}}{=}0$ (Figure~\ref{fig:geometry}b):
\begin{equation}
    T(\Delta z, S_{\mathrm{r}}) = e^{-1} + \sigma\!\left(e - \Delta z - e^{-1} S_{\mathrm{r}}\,\sigma(\Delta z - S_{\mathrm{r}})\right),
    \label{eq:ethermo}
\end{equation}
where $\sigma$ is the logistic sigmoid. Softmax over $\tilde{\mathbf{z}}/T$ gives System~1 triage probabilities $\mathbf{p}^{\mathrm{S1}}$ (ID-calibrated on Stage-1 dev, while $T{=}1$ softcapping gives the lowest tested OOD wECE; \S\ref{sec:ablations}), top choice $y^{\mathrm{S1}}$, and confidence $p_{\max}^{\mathrm{S1}}$. Positive $T$ preserves full rank order; $S_{\mathrm{r}}\sigma(\Delta z{-}S_{\mathrm{r}})$ is non-monotonic near $S_{\mathrm{r}}{\approx}\Delta z$, and fixed $\tau_z{=}e$ depends on logit scale (\appsec{sec:app_corpus}).

Third, free-energy margin $\Delta F = \max(0, \Delta z - S_{\mathrm{r}})$ partitions Figure~\ref{fig:geometry}c into \emph{Clear S1} ($\Delta F > \tau_z$), \emph{Noisy-Tail S1} ($\Delta z > \tau_z, \Delta F \le \tau_z$), and \emph{Ambiguous} ($\Delta z \le \tau_z$, $\tau_z{=}e$). \textbf{Joint Router} and \textbf{GO} exit without decoding when $\Delta z > \tau_z$; else \textbf{Joint Router} escalates iff $p_{\max}^{\mathrm{S1}} < \tau_p$ ($\tau_p{=}0.8$; $\Delta z/T - S_{\mathrm{r}} = \ln 4$ for tied rivals) while \textbf{GO} omits $\tau_p$.

Escalated System~2 decodes a CoT trace, reads terminal-slot logits over the $K$ option tokens post-trace, and applies debiasing, softcapping, and Eqs.~\eqref{eq:centering}--\eqref{eq:ethermo} for $(y^{\mathrm{S2}}, \mathbf{p}^{\mathrm{S2}})$; a post-\texttt{</think>} answer delimiter (\appsec{sec:app_repro}) overrides $y^{\mathrm{S2}}$ while evaluating $p^{\mathrm{S2}}_{y^{\mathrm{S2}}}$. A \textbf{Safety Ratchet} accepts $y^{\mathrm{S2}}$ iff $p^{\mathrm{S2}}_{y^{\mathrm{S2}}} \ge \max(\tau_r, 1/K)$ ($\tau_r{=}0.4$), else silently restores $y^{\mathrm{S1}}$ rather than abstaining (\S\ref{sec:limitations}).

\subsection{Data Decontamination and Three-Stage Post-Training}
\label{sec:training}

Our four scales initialize from Qwen3.5-Base. Training data are filtered against JevBench-231, Decision Index~0.3, and our evaluation suites via DSU twin clustering, normalized 13-gram matching, and MinHash-LSH (\appsec{sec:app_corpus}).

\textbf{Stage~1 (0.8B--9B): Single-Pass Alignment.} One-epoch single-pass distillation from Qwen3.8-Flash-Next using cross-entropy plus forward KL when teacher logits are available and margin loss otherwise \cite{qiu2026qwen38next}; 9B adds brief refinement.

\textbf{Stage~2 (0.8B--9B): CoT with Replay.} CoT traces mixed with ${\sim}25\%$ Stage-1 single-pass replay under response-masked cross-entropy (zero KL) and three depth-tiered learning-rate groups to protect mid-layer classification circuits.

\textbf{Stage~3 (4B \& 9B only): Interactive and Antithetic-Pair RLVR.} Multi-domain interactive RLVR runs at 4B and 9B only (not 0.8B/2B), followed by binary \textbf{ATP-RLVR} via clipped GRPO over coupled positive--negative twins ($\epsilon{=}0.2, \beta_{\mathrm{KL}}{=}0.04$) \cite{shao2024deepseekmath}, using four-quadrant rewards ($+1.0$ joint correct, $-2.0$ shortcut, $-2.5$ both wrong, $-1.5$ mixed) and mirror penalty $\lambda_{\mathrm{m}}\mathbb{E}[(m^+{+}m^-)^2]$ ($\lambda_{\mathrm{m}}{=}0.1$, $m^b{=}\tilde{z}^b_1{-}\tilde{z}^b_2$) updating only top blocks and head (\seeapptab[b]{tab:app_train_ablations}). Main inference evaluations reproduce from released BF16 checkpoints under greedy decoding within BF16 backend tolerance ($\le 0.125$\,nats; \appsec{sec:app_repro}); proprietary schedules remain high-level.

\section{Experiments}
\label{sec:experiments}

We evaluate Tractus on nine benchmarks (256K capacity; 64K--256K untested), diagnostic suites, the Commercial Suite, and Decision Index~0.3, covering setup (\S\ref{sec:protocol}), the 4B threshold (\S\ref{sec:unified_panorama}), ablations (\S\ref{sec:ablations}), and case studies (\S\ref{sec:qualitative}).

\subsection{Experimental Setup}
\label{sec:protocol}

\begin{table*}[t]
\centering
\small
\renewcommand{\arraystretch}{0.89}
\setlength{\tabcolsep}{2.4pt}
\begin{tabular*}{\textwidth}{@{\extracolsep{\fill}} lcccccccc @{}}
\toprule
& \multicolumn{2}{c}{\textbf{0.8B}} & \multicolumn{2}{c}{\textbf{2B}} & \multicolumn{2}{c}{\textbf{4B}} & \multicolumn{2}{c}{\textbf{9B}} \\
\cmidrule(lr){2-3}\cmidrule(lr){4-5}\cmidrule(lr){6-7}\cmidrule(lr){8-9}
\textbf{Benchmark (Metric, Count)} & \textbf{Dec.} & \textbf{Dual} & \textbf{Dec.} & \textbf{Dual} & \textbf{Dec.} & \textbf{Dual} & \textbf{Dec.} & \textbf{Dual} \\
\midrule
JevBench-231\textsuperscript{*} (Acc, 231) & \textbf{72.73\%} & 70.13\% & \textbf{69.26\%} & 65.37\% & 76.19\% & \textbf{88.31\%} & 79.22\% & \textbf{90.91\%} \\
Global-MMLU-EN (Acc, 400) & \textbf{49.00\%} & 47.50\% & \textbf{60.00\%} & 59.00\% & 79.25\% & \textbf{82.75\%} & 82.25\% & \textbf{85.75\%} \\
Global-MMLU-Lite (Acc, 1,200) & \textbf{39.92\%} & 38.92\% & 48.42\% & \textbf{49.00\%} & 67.08\% & \textbf{77.33\%} & 73.50\% & \textbf{81.67\%} \\
Global-MMLU-Non-EN (Acc, 2,200) & 38.23\% & \textbf{39.14\%} & \textbf{43.77\%} & 43.59\% & 60.14\% & \textbf{72.55\%} & 66.86\% & \textbf{79.00\%} \\
When2Call (Acc, 2,000) & \textbf{39.85\%} & 38.85\% & \textbf{58.75\%} & 57.05\% & 65.20\% & \textbf{80.30\%} & 76.05\% & \textbf{82.20\%} \\
Amazon ESCI (Macro-F1, 2,000) & 26.66\% & \textbf{27.19\%} & \textbf{29.54\%} & 28.02\% & 41.07\% & \textbf{46.39\%} & 37.68\% & \textbf{46.26\%} \\
Amazon ESCI\textsuperscript{$\dagger$} (Top-1, 2,000) & \textbf{52.00\%} & 48.25\% & \textbf{55.75\%} & 52.05\% & \textbf{61.60\%} & 61.10\% & 60.50\% & \textbf{62.45\%} \\
Mind2Web (Acc, 276) & 63.77\% & 63.77\% & 74.28\% & \textbf{75.36\%} & \textbf{86.96\%} & 84.78\% & 86.23\% & \textbf{87.32\%} \\
PubMedQA (Acc, 1,000) & 67.50\% & \textbf{67.80\%} & \textbf{72.20\%} & 67.60\% & \textbf{69.20\%} & 69.00\% & \textbf{75.60\%} & 73.30\% \\
New Yorker Caption (Acc, 528) & \textbf{32.95\%} & 31.82\% & \textbf{45.27\%} & 43.75\% & 49.43\% & \textbf{52.65\%} & 65.72\% & \textbf{67.05\%} \\
\midrule
\textbf{9-Task Macro Mean} (9 tasks) & \textbf{47.84\%} & 47.23\% & \textbf{55.72\%} & 54.30\% & 66.06\% & \textbf{72.67\%} & 71.46\% & \textbf{77.05\%} \\
\textbf{Pooled 9-Task Acc.} (Top-1, 9,835) & \textbf{46.22\%} & 45.19\% & \textbf{54.91\%} & 53.19\% & 64.57\% & \textbf{72.08\%} & 70.53\% & \textbf{76.18\%} \\
\bottomrule
\end{tabular*}
\caption{Nine-benchmark BF16 evaluation (256K context capacity, $9{,}835$ items per scale) comparing internal \tractusdec\ (Dec.) against dual-mode \tractus\ (Dual); \textbf{bold} marks the better mode per scale. \textsuperscript{*}2B--9B JevBench-231 uses an 8K cap (Figure~\ref{fig:scaling_panorama}a shows the 4K run). \textsuperscript{$\dagger$}Macro Mean uses ESCI Macro-F1; Pooled uses Top-1.}
\label{tab:unified_panorama}
\end{table*}

\begin{figure*}[t]
    \centering
    \includegraphics[width=0.95\textwidth]{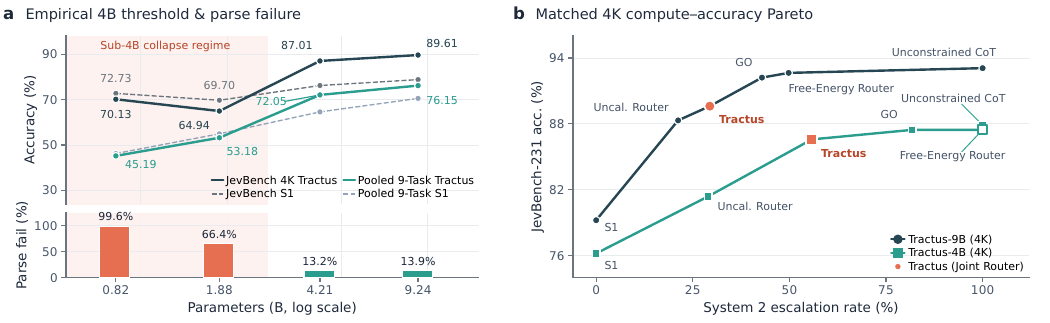}
    \caption{Empirical 4B threshold (co-occurring with Stage-3 RLVR and 32-layer depth) and matched 4K Pareto frontier. (a)~Pooled 9-Task accuracy, batched vLLM 4K JevBench-231 accuracy ($87.01\%$ at 4B), and System~2 parse-failure rates across $9{,}835$ items. (b)~Single-stream Transformers 4K routing ablation, where 4B Tractus reaches $86.58\%$ (a one-item cross-backend difference from (a); see \appsec{sec:app_repro}).}
    \label{fig:scaling_panorama}
\end{figure*}

\paragraph{Primary Benchmarks.}
Nine BF16 benchmarks run at 256K context capacity (64K--256K untested), covering JevBench-231, Global-MMLU EN/Lite/Non-EN \cite{singh2025globalmmlu}, When2Call \cite{ross2025when2call}, Amazon ESCI \cite{reddy2022esci}, Mind2Web \cite{deng2023mind2web}, PubMedQA \cite{jin2019pubmedqa}, and New Yorker Caption \cite{hessel2023newyorker}. Over $9{,}835$ queries/scale, Table~\ref{tab:unified_panorama} compares internal single-pass vs.\ dual-mode runs (\S\ref{sec:benchmark} for external baselines), reporting \textbf{Pooled 9-Task Accuracy} and unweighted \textbf{9-Task Macro Mean} (ESCI Macro-F1). Decision Index~0.3\footnote{\raggedright JevBench: \url{https://github.com/fstandhartinger/jevbench}; Decision Index~0.3: \url{https://huggingface.co/spaces/multimodalart/jev-decision-index}.} reports Balanced Skill, Balanced Raw, and Breadth across 18 public baselines, including vLLM-SR Decision and Drex \cite{liu2026decision1,dhankhar2026drex}.

\paragraph{Diagnostic Suites.}
Single-pass calibration runs on $33{,}803$ held-out 0.8B Stage-1 dev queries (Table~\ref{tab:calibration}a) and seven full OOD splits totaling $19{,}887$ queries (When2Call, Global-MMLU-Lite, ESCI, PubMedQA, New Yorker, Mind2Web, JevBench-231; Table~\ref{tab:calibration}b, \apptab{tab:app_stage_c}) via weighted accuracy, 15-bin wECE, and CycFlip. Routing uses the 550-item \textbf{Five-Domain Suite} (110 items per domain across MMLU-EN, MMLU-Non-EN, Mind2Web, PubMedQA, and When2Call) and the 202-item \textbf{A/V Subset} combining MMAU and unsubtitled Video-MME \cite{sakshi2024mmau,fu2025videomme}. Retention uses a 128-item 2B web holdout, the 1,218-item \textbf{General NLU} suite, and the 12-task \textbf{Web-Trajectory} suite.

\paragraph{Backends and Baselines.}
BF16 runs on RTX PRO 6000 Blackwell GPUs use greedy decoding (zero sampling variance) via cycle-halted single-stream Transformers and batched vLLM on high-volume splits \cite{wolf2020transformers,kwon2023vllm}. The 831-item \textbf{Commercial Suite} tests three prompts per Tractus mode vs.\ single-prompt Qwen3.5 and external full-split DataCamp APIs.\footnote{\raggedright\url{https://www.datacamp.com/blog/gemini-decision-1}} Same-checkpoint variants include \textbf{Uncalibrated Router} ($p_{\mathrm{raw}}{<}0.8$), \textbf{Margin-Only Thermostat} ($S_{\mathrm{r}}{=}0$), \textbf{GO}, \textbf{Free-Energy Router} ($\Delta F{\le}e$), \textbf{No Safety Ratchet}, and \textbf{Unconstrained CoT}.

\subsection{Main Results and the Empirical 4B Threshold}
\label{sec:unified_panorama}

Table~\ref{tab:unified_panorama} and Figure~\ref{fig:scaling_panorama}a show an empirical 4B threshold in our family under greedy decoding: aggregate \tractusdec\ accuracy scales monotonically from 0.8B to 9B, whereas aggregate dual-mode gains turn positive at 4B ($+7.51$\,pp Pooled at 4B; $+5.65$\,pp at 9B). Because Stage-3 RLVR (4B/9B only), 32- vs.\ 24-layer depth, and greedy loops co-occur at 2B$\to$4B, this threshold reflects our pipeline rather than parameter count alone. Major dual-mode gains concentrate on JevBench-231 alongside MMLU-Non-EN, MMLU-Lite, and When2Call, with the latter three accounting for the vast majority of the 4B and 9B pooled improvement. Table~\ref{tab:unified_panorama} caps 2B--9B JevBench-231 at 8K tokens, whereas Figure~\ref{fig:scaling_panorama}a reports the uniform 4K run (\apptab[a]{tab:app_4k_fig4}).

\begin{table*}[t]
\centering
\small
\renewcommand{\arraystretch}{0.86}
\setlength{\tabcolsep}{2.5pt}
\begin{tabular*}{\textwidth}{@{\extracolsep{\fill}} lcccccccc @{}}
\toprule
\multicolumn{9}{@{}l}{\textbf{(a) Stepwise 4B Component Ablation on JevBench-231 (8K Cap; All Steps Achieve 48/48 Easy)}} \\
\midrule
Ablation Step & ECE $\downarrow$ & S2\% & Tok & Tok Cut & Std (72) & Hard (111) & Total Acc & Neg.\ Flip \\
\midrule
1. Raw S1 ($T=1$) & 11.48\% & 0.0\% & 0 & $-$100\% & 67/72 & 61/111 & 76.19\% (176) & 0.00\% \\
2. $+$ Softcap \& Centering ($T=1$) & 8.60\% & 0.0\% & 0 & $-$100\% & 67/72 & 61/111 & 76.19\% (176) & 0.00\% \\
3. $+$ $e$-Thermostat & 8.29\% & 0.0\% & 0 & $-$100\% & 67/72 & 61/111 & 76.19\% (176) & 0.00\% \\
4. $+$ Joint Router, No Safety Ratchet & 8.16\% & 56.3\% & 1,275 & $-$12.7\% & \textbf{69/72} & 86/111 & 87.88\% (203) & 3.03\% \\
5. $+$ Safety Ratchet (\textbf{Tractus}) & \textbf{7.67\%} & 56.3\% & 1,275 & $-$12.7\% & \textbf{69/72} & 87/111 & \textbf{88.31\% (204)} & \textbf{2.60\%} \\
6. Unconstrained CoT & 9.48\% & 100.0\% & 1,460 & 0.0\% & 68/72 & \textbf{88/111} & \textbf{88.31\% (204)} & 4.33\% \\
\bottomrule
\end{tabular*}

\vspace{2pt}
\setlength{\tabcolsep}{2.2pt}
\begin{tabular*}{\textwidth}{@{\extracolsep{\fill}} lcccccccc @{}}
\toprule
\multicolumn{9}{@{}l}{\textbf{(b) Cross-Scale Routing, Safety Ratchet, End-to-End Token Cost, and Overthinking Summary}} \\
\midrule
Suite \& Scale (Cap) & S1 Acc & \thdr{No Safety\\Ratchet} & Tractus & S2\% & Tok\,(D/U) & Tok Cut & \thdr{Unconstrained\\CoT} & vs.\ S1 (pp) \\
\midrule
JevBench 9B (8K) & 79.22\% & 90.91\%\,(6.49\%) & \textbf{90.91\%}\,(6.78\%) & 29.0\% & 751/1,214 & $-$38.1\% & 94.37\% & $+$11.69 \\
JevBench 4B (8K) & 76.19\% & 87.88\%\,(8.16\%) & \textbf{88.31\%}\,(7.67\%) & 56.3\% & 1,275/1,460 & $-$12.7\% & 88.31\% & $+$12.12 \\
JevBench 2B (8K) & 69.26\% & 62.77\%\,(12.24\%) & 65.37\%\,(11.96\%) & 45.0\% & 823/1,067 & $-$22.9\% & 61.47\% & $-$3.89 \\
JevBench 0.8B (4K) & 72.73\% & 69.70\%\,(15.91\%) & 70.13\%\,(16.70\%) & 29.0\% & 5/16 & $-$68.8\% & 70.13\% & $-$2.60 \\
\midrule
Five-Domain 9B (4K) & 79.45\% & 79.27\%\,(19.63\%) & 79.27\%\,(19.55\%) & 43.3\% & 516/919 & $-$43.9\% & 79.09\% & $-$0.18 \\
Five-Domain 4B (4K) & 74.00\% & 78.18\%\,(21.36\%) & \textbf{78.55\%}\,(21.48\%) & 77.8\% & 886/1,019 & $-$13.1\% & 78.18\% & $+$4.55 \\
Five-Domain 2B (4K) & 64.18\% & 58.55\%\,(12.11\%) & 62.00\%\,(11.21\%) & 73.8\% & 410/446 & $-$8.1\% & 58.18\% & $-$2.18 \\
Five-Domain 0.8B (4K) & 50.73\% & 53.45\%\,(28.44\%) & \textbf{53.45\%}\,(28.42\%) & 42.0\% & 8/17 & $-$52.9\% & 52.00\% & $+$2.72 \\
\midrule
A/V Subset 4B (4K) & 63.86\% & -- & 58.42\%\,(19.22\%) & 76.2\% & 1,762/1,925 & $-$8.5\% & 57.92\% & $-$5.44 \\
A/V Subset 9B (4K) & 64.36\% & -- & 62.87\%\,(18.88\%) & 53.0\% & 930/1,334 & $-$30.3\% & 61.39\% & $-$1.49 \\
\bottomrule
\end{tabular*}
\caption{Thermostasis component ablation and cross-scale summary. In (a), \textbf{bold} marks the best result per column (lowest Neg.\ Flip among System~2 rows). In (b), \textbf{bold} marks Tractus outperforming System~1; No Safety Ratchet and Tractus report accuracy (15-bin ECE in parentheses), Tok\,(D/U) lists dual-mode and unconstrained mean tokens (0.8B's 5/16 tokens reflect single-stream repetition-cycle halting vs.\ 1,605.8 unhalted vLLM tokens in \apptab[a]{tab:app_4k_fig4}), Tok Cut gives reduction vs.\ Unconstrained CoT, `--' marks No Safety Ratchet not separately ablated on A/V Subset, and vs.\ S1 is Tractus minus S1.}
\label{tab:ablation}
\end{table*}

Sub-4B dual-mode losses stem mainly from decoding failures rather than prefill scoring. Under the 4K cap (\apptab[a]{tab:app_4k_fig4}, Figure~\ref{fig:scaling_panorama}a), 0.8B and 2B frequently loop or hit the cap without closing traces, whereas parse failures drop sharply at 4B and 9B; thus \tractusdec\ is preferable below 4B, aside from an isolated cycle-halted 0.8B Five-Domain gain (Table~\ref{tab:ablation}b).

\subsection{Ablation Studies and Calibration Analysis}
\label{sec:ablations}

\begin{table}[t]
\centering
\small
\renewcommand{\arraystretch}{0.92}
\setlength{\tabcolsep}{2.4pt}
\begin{tabular*}{\columnwidth}{@{\extracolsep{\fill}} lcccc @{}}
\toprule
\multicolumn{5}{@{}l}{\textbf{(a) Validation Calibration (0.8B Stage-1 Dev., 33,803 items)}} \\
\midrule
Method & Fitted & ECE $\downarrow$ & Brier $\downarrow$ & NLL $\downarrow$ \\
\midrule
Uncalibrated ($T=1.0$) & 0 & 2.79\% & 0.4037 & 0.7189 \\
Log-Cardinality Scaling & 2 & 3.97\% & 0.4072 & 0.7321 \\
Static Temp. ($T=0.84$) & 1 & 2.14\% & 0.4035 & 0.7176 \\
3-Param Oracle (In-Dist.) & 3 & 0.75\% & 0.4030 & 0.7127 \\
\textbf{$e$-Thermostat (Ours)} & \textbf{0} & \textbf{1.16\%} & \textbf{0.4034} & \textbf{0.7164} \\
\midrule
\multicolumn{5}{@{}l}{\textbf{(b) OOD Full-Split Single-Pass (19,887 items, zero System~2)}} \\
\midrule
& \multicolumn{2}{c}{\textbf{4B}} & \multicolumn{2}{c}{\textbf{9B}} \\
\cmidrule(lr){2-3}\cmidrule(lr){4-5}
Setting & wAcc & wECE & wAcc & wECE \\
\midrule
Raw Logits & 62.66\% & 10.62\% & 68.09\% & 13.77\% \\
Softcap \& Center ($T{=}1$) & 62.66\% & \textbf{8.28\%} & 68.09\% & \textbf{6.85\%} \\
$+$ 2D $e$-Thermostat & 62.66\% & 10.80\% & 68.09\% & 8.52\% \\
\bottomrule
\end{tabular*}
\caption{(a)~0.8B Stage-1 validation calibration across 89 domains (all five preserve $65.15\%$ top-1 accuracy), and (b)~OOD single-pass calibration at $\mathbf{z}_0{=}\mathbf{0}$, $C{=}30$ (see \apptab{tab:app_stage_c}). \textbf{Bold} marks the best unfitted method in (a) and lowest wECE per scale in (b).}
\label{tab:calibration}
\end{table}

Stepwise 4B ablations (Table~\ref{tab:ablation}a, $\mathbf{z}_0{=}\mathbf{0}$) show softcapping ($C{=}30$) drives most single-pass JevBench ECE reduction (cross-backend shift vs.\ \apptab[b]{tab:app_4k_fig4}; \appsec{sec:app_repro}). Adding the Joint Router recovers most of the gap to Unconstrained CoT, and the Safety Ratchet recovers one additional Hard item to match Unconstrained CoT with $12.7\%$ fewer tokens and fewer negative flips. Across scales (Table~\ref{tab:ablation}b), 9B cuts tokens substantially on both JevBench ($-38.1\%$, trailing Unconstrained CoT by $3.46$\,pp) and Five-Domain ($-43.9\%$, matching S1 accuracy). The Safety Ratchet is largely inert at 4B--9B while providing its clearest protection at 2B by reverting failed rollouts on both suites. Training ablations (\apptab{tab:app_train_ablations}) indicate that tiered Stage-2 learning rates avoid the uniform-LR 2B holdout drop, while 9B ATP-RLVR sharply reduces phishing logit bias without General NLU or Web-Trajectory regressions.

Table~\ref{tab:calibration} isolates single-pass calibration. On 0.8B Stage-1 in-distribution validation (Table~\ref{tab:calibration}a; no $T{=}1$ softcap row logged), Thermostasis reduces 15-bin ECE by $58.5\%$ relative to raw logits with modest Brier and NLL gains, whereas log-cardinality scaling degrades ECE; thresholding at $\Delta z{=}e$ cleanly separates high-margin queries (near-ceiling accuracy) from lower-margin ambiguous ones. Across the seven OOD splits (Table~\ref{tab:calibration}b, \apptab{tab:app_stage_c}, $\mathbf{z}_0{=}\mathbf{0}$), unit-temperature softcapping ($T{=}1$) yields the lowest tested single-pass wECE at both 4B and 9B, whereas adding the 2D $e$-thermostat \emph{raises} OOD wECE over $T{=}1$ (exceeding raw logits at 4B), and dual-mode escalation also raises ECE on Five-Domain (\apptab[b]{tab:app_4k_fig4}) and the Commercial Suite (Table~\ref{tab:decision_index}b). Rank-preserving single-pass Thermostasis leaves cyclic flip rates invariant ($22.3\%\text{--}42.4\%$). Together, these results indicate that softcapping drives OOD calibration gains while the $e$-thermostat serves mainly for triage.

\subsection{Case Studies and Overthinking Analysis}
\label{sec:qualitative}

\begin{figure}[t]
    \centering
    \includegraphics[width=0.96\columnwidth]{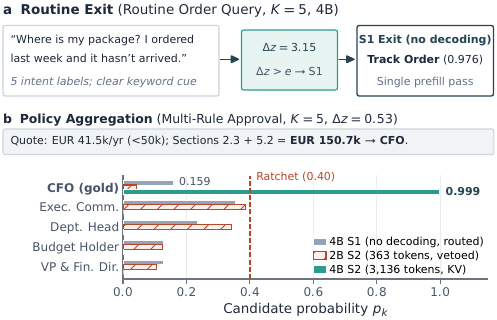}
    \caption{Qualitative JevBench-231 examples ($K=5$). (a)~A routine inquiry exits without decoding. (b)~A multi-rule approval query triggers 4B System~2 deliberation.}
    \label{fig:qualitative_case}
\end{figure}

Figure~\ref{fig:qualitative_case} illustrates dual-mode routing on two 5-option JevBench-231 queries. In panel a, a routine order inquiry yields $\Delta z = 3.15 > e$ and exits without decoding at $0.976$ confidence on \textit{Track Order} in a single prefill pass.
In panel b, an annual quote of EUR~$41.5$k falls below a EUR~$50$k cap, but Sections~2.3 and~5.2 sum to EUR~$150.7$k and require \textit{CFO} sign-off. At prefill, 4B System~1 yields a narrow margin ($\Delta z = 0.53 \le e$) favoring \textit{Exec.\ Comm.} at $0.354$ over \textit{CFO} at $0.159$ and triggers escalation. The 2B model loops for $363$\,tokens and peaks at $0.388$ on \textit{Exec.\ Comm.} and $0.042$ on \textit{CFO}, below the $0.4$ ratchet floor so System~1 is restored; 4B resolves the policy terms in $3{,}136$\,tokens at $0.999$ for \textit{CFO}; 9B reaches $1.000$.

Routing ablations (Figure~\ref{fig:scaling_panorama}b, \apptab[b]{tab:app_4k_fig4}) compare single operating points without threshold-swept AURC curves. On Five-Domain, the Uncalibrated Router Pareto-dominates Tractus across accuracy, wECE, and token cost at both 4B and 9B, while escalating less and trailing Tractus on 4K JevBench. The Free-Energy Router escalates high-accuracy Noisy-Tail items, pushing 4B to full escalation on both suites, while GO likewise drives heavy 4B escalation without improving Five-Domain accuracy. System~2 degrades PubMedQA across 2B--9B (Table~\ref{tab:unified_panorama}), and Unconstrained CoT degrades the A/V Subset (Table~\ref{tab:ablation}b).

\section{Decision Index 0.3 and Commercial Baselines}
\label{sec:benchmark}

\begin{table}[t]
\centering
\small
\renewcommand{\arraystretch}{0.83}
\setlength{\tabcolsep}{2.2pt}
\begin{tabular*}{\columnwidth}{@{\extracolsep{\fill}} llccc @{}}
\toprule
\multicolumn{5}{@{}l}{\textbf{(a) Decision Index 0.3 Top-3 per Tier (2,172 items)}} \\
\midrule
Tier & Model & Size & Skill $\uparrow$ & Raw \\
\midrule
\multirow{3}{*}{9B}
& \textbf{Tractus-dec.-9B} & 9.24B & \textbf{46.03} & 59.32\% \\
& vLLM-SR Lux 9B & 9.65B & 45.97 & 59.93\% \\
& JevK5 9B v0.3.3 & 9.65B & 44.89 & \textbf{60.38\%} \\
\midrule
\multirow{3}{*}{4B}
& \textbf{Tractus-dec.-4B} & 4.21B & \textbf{41.31} & 55.56\% \\
& vLLM-SR Nox 4B & 4.21B & 40.75 & \textbf{58.08\%} \\
& Hopper (G) 1.2 & 4.66B & 40.17 & 56.00\% \\
\midrule
\multirow{3}{*}{2B}
& \textbf{Tractus-dec.-2B} & 1.88B & \textbf{29.27} & 46.83\% \\
& this-that 1.2 (2B) & 1.88B & 24.85 & 45.76\% \\
& Decider 2B & 2.27B & 23.65 & \textbf{47.01\%} \\
\midrule
\multirow{3}{*}{Sub-2B}
& \textbf{Tractus-dec.-0.8B} & 0.82B & \textbf{18.07} & 37.66\% \\
& Bosun v3.1 1.7B & 1.72B & 16.54 & \textbf{41.06\%} \\
& JPT-0.8B & 0.87B & 15.46 & 39.36\% \\
\bottomrule
\end{tabular*}

\vspace{1pt}
\setlength{\tabcolsep}{1.0pt}
\begin{tabular*}{\columnwidth}{@{\extracolsep{\fill}} lccccc @{}}
\toprule
\multicolumn{6}{@{}l}{\textbf{(b) 831-Task Suite (3-Prompt vs.\ 1-Prompt)}} \\
\midrule
System & JB $\uparrow$ & ECE $\downarrow$ & M2W & W2C & Med/Mean\,(s) \\
\midrule
\textbf{Tractus-d.-9B} & 80.5\% & \textbf{2.45\%} & \textbf{88.0\%} & 75.0\% & \textbf{0.14/0.18}\textsuperscript{*} \\
\textbf{Tractus-9B} & 88.7\% & 5.28\% & 85.0\% & \textbf{84.0\%} & 0.69/7.65 \\
\textbf{Tractus-9B GO} & 92.2\% & 2.80\% & 85.0\% & \textbf{84.0\%} & 7.22/10.1 \\
\midrule
Qwen3.5-9B & \textbf{93.1\%} & -- & 79.0\% & \textbf{84.0\%} & 26.0/34.0 \\
Qwen3.5-4B & 89.2\% & -- & 78.0\% & 81.0\% & 14.4/22.1 \\
\midrule
\multicolumn{6}{@{}l}{\textit{External Full-Split APIs (DataCamp, non-matched)}} \\
Gemini Dec.\ 1 & 92.2\% & 2.50\% & 86.2\% & 79.9\% & -- \\
Jev 1.14 & 90.0\% & 4.94\% & 86.8\% & 82.0\% & -- \\
GPT-6 Luna & 90.9\% & 7.21\% & 81.8\% & 70.4\% & -- \\
Flash-Lite & 86.6\% & 21.04\% & 78.0\% & 60.9\% & -- \\
\bottomrule
\end{tabular*}
\caption{(a)~Decision Index~0.3 top-3/tier (2,172 items). (b)~831-task suite (fixed order; Tractus-9B trails Qwen3.5-9B on 4/7 tasks---JB, ESCI, PubMedQA, New Yorker; all 7 in \apptab[a]{tab:commercial_suite}): best-of-3 Tractus (`d.': decision) vs.\ 1-prompt Qwen3.5 Thinking; 10-bin unfitted ECE (non-comparable to \S\ref{sec:experiments}'s 15-bin); Med/Mean: $n{=}70$ GPU latency (\textsuperscript{*}1-prompt: $47.4$/$69.2$\,ms). \textbf{Bold}: best in tier (a) / local (b).}
\label{tab:decision_index}
\end{table}

Table~\ref{tab:decision_index}a evaluates \tractusdec\ on Decision Index~0.3 (2,172 items; \seeapptab{tab:app_decision_index}), where it ranks first among compared baselines in Balanced Skill across all four tiers (by a narrow $+0.06$ margin at 9B using two priors without a $\mathbf{z}_0{=}\mathbf{0}$ ablation), while trailing specialized baselines on Balanced Raw (5th at 9B) and 9B Breadth.

Table~\ref{tab:decision_index}b and \apptab{tab:commercial_suite} report an exploratory 831-task comparison (fixed option order; $N{=}100$ splits, $\text{SE}{\approx}3\text{--}5$\,pp) with four protocol asymmetries: best-of-3 Tractus prompts vs.\ single-prompt Qwen3.5 (with fixed-order prompts contributing to the JevBench difference from Table~\ref{tab:unified_panorama}), full-split DataCamp external APIs, closed-menu logit recovery ($0$ Tractus failures vs.\ 9B's 4K delimiter failures in \apptab[a]{tab:app_4k_fig4} and Qwen3.5 extraction failures), and sequential latency on an $n{=}70$ probe. Across the seven shared tasks (\seeapptab[a]{tab:commercial_suite}), \tractus-9B trails Qwen3.5-9B Thinking on JevBench, ESCI, PubMedQA, and New Yorker (and \tractusdec-9B on the last three plus Mind2Web), while leading Qwen3.5-9B on Mind2Web and MMLU-Lite and tying When2Call. Single-pass \tractusdec-9B runs at $47.4$\,ms median per prompt ($2.45\%$ 10-bin ECE); \tractus-9B resolves over half of queries ($479/831$) without decoding, cutting median and tail latency substantially vs.\ Qwen3.5-9B Thinking (Table~\ref{tab:decision_index}b), while GO reaches $92.2\%$ on JevBench at higher latency.

\section{Conclusion and Limitations}
\label{sec:conclusion}
\label{sec:limitations}

We introduced \textbf{Tractus} (0.8B--9B), uniting single-pass scoring and prefix-reusing CoT under Thermostasis (zero fitted temperatures, five constants, two optional 9B DI~0.3 priors). Within our family under greedy decoding, adaptive deliberation emerges at 4B, ablations separate softcapping OOD calibration from thermostat triage, \tractusdec\ leads DI~0.3 Balanced Skill across all four tiers (trailing Balanced Raw), and 3-prompt Tractus-9B resolves over half of the 831-task suite without decoding at lower latency than 1-prompt Qwen3.5-9B Thinking with zero logit-recovery failures.

\textbf{Limitations.} The 4B threshold co-occurs with Stage-3 RLVR (4B/9B), 32-layer depth, and greedy loops, favoring \tractusdec\ below 4B. OOD at 4B--9B, the 2D $e$-thermostat raises wECE over $T{=}1$ softcapping, leaves cyclic flip rates invariant, and trails Uncalibrated Router on Five-Domain; AURC curves and external cascade/budget-CoT baselines remain open. The Safety Ratchet is largely inert at 4B--9B, unstratified in $K$, restoring System~1 without abstaining, while the two 9B DI~0.3 priors are base-rate sensitive without a $\mathbf{z}_0{=}\mathbf{0}$ ablation. Open items include 64K--256K tests, KV-reuse timing ablations, non-English System~1 gaps, native audio/video, and Stage-3 ID calibration (Table~\ref{tab:calibration}a audits 0.8B Stage-1; \appsec{sec:app_repro}).

\bibliography{references}

\begin{thebibliography}{50}
\providecommand{\natexlab}[1]{#1}

\bibitem[{Alomrani et~al.(2025)Alomrani, Zhang, Li, Sun, Pal, Zhang, Hu,
  Ajwani, Valkanas, Karimi, Cheng, Wang, Liao, Huang, Wang, Hao, and
  Coates}]{alomrani2025budget}
Alomrani, M.~A.; Zhang, Y.; Li, D.; Sun, Q.; Pal, S.; Zhang, Z.; Hu, Y.;
  Ajwani, R.~D.; Valkanas, A.; Karimi, R.; Cheng, P.; Wang, Y.; Liao, P.;
  Huang, H.; Wang, B.; Hao, J.; and Coates, M. 2025.
\newblock Reasoning on a Budget: A Survey of Adaptive and Controllable
  Test-Time Compute in {LLMs}.
\newblock \emph{arXiv preprint arXiv:2507.02076}.

\bibitem[{Banino, Balaguer, and Blundell(2021)}]{banino2021pondernet}
Banino, A.; Balaguer, J.; and Blundell, C. 2021.
\newblock {PonderNet}: Learning to Ponder.
\newblock \emph{arXiv preprint arXiv:2107.05407}.

\bibitem[{Chen, Zaharia, and Zou(2024)}]{chen2023frugalgpt}
Chen, L.; Zaharia, M.; and Zou, J. 2024.
\newblock {FrugalGPT}: How to Use Large Language Models While Reducing Cost and
  Improving Performance.
\newblock \emph{Trans. Mach. Learn. Res.}

\bibitem[{Deng et~al.(2023)Deng, Gu, Zheng, Chen, Stevens, Wang, Sun, and
  Su}]{deng2023mind2web}
Deng, X.; Gu, Y.; Zheng, B.; Chen, S.; Stevens, S.; Wang, B.; Sun, H.; and Su,
  Y. 2023.
\newblock {Mind2Web}: Towards a Generalist Agent for the Web.
\newblock In \emph{Proc. NeurIPS}, 28091--28114.

\bibitem[{Dhankhar et~al.(2026)Dhankhar, Amantay, Datbayev, Akhmetov, Shah,
  Ubaidullayev, Saparkhan, Baha, and Saparov}]{dhankhar2026drex}
Dhankhar, N.; Amantay, A.; Datbayev, Z.; Akhmetov, D.; Shah, Y.; Ubaidullayev,
  N.; Saparkhan, R.; Baha, D.; and Saparov, A. 2026.
\newblock Towards Foundation Decision Models: {Drex}.
\newblock \emph{arXiv preprint arXiv:2611.07854}.

\bibitem[{Duan et~al.(2026)Duan, Lin, Wang, Xiang, Wu, Zhang, Yan, Wang, and
  Wei}]{duan2026jeval}
Duan, F.; Lin, J.; Wang, J.; Xiang, J.; Wu, J.; Zhang, X.; Yan, H.; Wang, S.;
  and Wei, Z. 2026.
\newblock General Decision Models: Benchmarking and Insights Beyond {Jev}.
\newblock \emph{arXiv preprint arXiv:2610.03935}.

\bibitem[{Fatemi et~al.(2026)Fatemi, Shi, Mueller, and Ma}]{fatemi2026gps}
Fatemi, M.~S.; Shi, Z.; Mueller, K.; and Ma, T. 2026.
\newblock Guided Permutation Scoring ({GPS}): Option-Order Permutation for
  Multiple-Choice Question Answering.
\newblock In \emph{Findings of EMNLP}, 27294--27313.

\bibitem[{Fu et~al.(2025)Fu, Dai, Luo, Li, Ren, Zhang, Wang, Zhou, Shen, Zhang,
  Chen, Li, Lin, Zhao, Li, Xu, Zheng, Chen, Shan, He, and Sun}]{fu2025videomme}
Fu, C.; Dai, Y.; Luo, Y.; Li, L.; Ren, S.; Zhang, R.; Wang, Z.; Zhou, C.; Shen,
  Y.; Zhang, M.; Chen, P.; Li, Y.; Lin, S.; Zhao, S.; Li, K.; Xu, T.; Zheng,
  X.; Chen, E.; Shan, C.; He, R.; and Sun, X. 2025.
\newblock {Video-MME}: The First-Ever Comprehensive Evaluation Benchmark of
  Multi-modal {LLMs} in Video Analysis.
\newblock In \emph{Proc. CVPR}, 24108--24118.

\bibitem[{{Gemma Team} et~al.(2024){Gemma Team}, Riviere, Pathak, Sessa,
  Hardin, Bhupatiraju, Hussenot, Mesnard, Shahriari, Ram{\'e}
  et~al.}]{gemma2024gemma2}
{Gemma Team}; Riviere, M.; Pathak, S.; Sessa, P.~G.; Hardin, C.; Bhupatiraju,
  S.; Hussenot, L.; Mesnard, T.; Shahriari, B.; Ram{\'e}, A.; et~al. 2024.
\newblock {Gemma 2}: Improving Open Language Models at a Practical Size.
\newblock \emph{arXiv preprint arXiv:2408.00118}.

\bibitem[{Ghosal et~al.(2025)Ghosal, Chakraborty, Reddy, Lu, Wang, Manocha,
  Huang, Ghavamzadeh, and Bedi}]{ghosal2025mirage}
Ghosal, S.~S.; Chakraborty, S.; Reddy, A.; Lu, Y.; Wang, M.; Manocha, D.;
  Huang, F.; Ghavamzadeh, M.; and Bedi, A.~S. 2025.
\newblock Does Thinking More Always Help? Mirage of Test-Time Scaling in
  Reasoning Models.
\newblock In \emph{Proc. NeurIPS}, 191359--191386.

\bibitem[{Guo et~al.(2017)Guo, Pleiss, Sun, and
  Weinberger}]{guo2017calibration}
Guo, C.; Pleiss, G.; Sun, Y.; and Weinberger, K.~Q. 2017.
\newblock On Calibration of Modern Neural Networks.
\newblock In \emph{Proc. ICML}, 1321--1330.

\bibitem[{Guo et~al.(2025)Guo, Yang, Zhang, Song, Wang, Zhu, Xu, Zhang, Ma, Bi
  et~al.}]{guo2025deepseekr1}
Guo, D.; Yang, D.; Zhang, H.; Song, J.; Wang, P.; Zhu, Q.; Xu, R.; Zhang, R.;
  Ma, S.; Bi, X.; et~al. 2025.
\newblock {DeepSeek-R1} Incentivizes Reasoning in {LLMs} through Reinforcement
  Learning.
\newblock \emph{Nature}, 645: 633--638.

\bibitem[{Hanna and Feng(2026)}]{hanna2026decoupling}
Hanna, K.; and Feng, C. 2026.
\newblock {A, B, C}, or Stable? Accuracy and Order Sensitivity Diverge in {LLM}
  Multiple-Choice Evaluation.
\newblock \emph{arXiv preprint arXiv:2608.11947}.

\bibitem[{Hassid et~al.(2025)Hassid, Synnaeve, Adi, and
  Schwartz}]{hassid2025shorter}
Hassid, M.; Synnaeve, G.; Adi, Y.; and Schwartz, R. 2025.
\newblock Don't Overthink It. Preferring Shorter Thinking Chains for Improved
  {LLM} Reasoning.
\newblock \emph{arXiv preprint arXiv:2505.17813}.

\bibitem[{Hessel et~al.(2023)Hessel, Marasovi{\'c}, Hwang, Lee, Da, Zellers,
  Mankoff, and Choi}]{hessel2023newyorker}
Hessel, J.; Marasovi{\'c}, A.; Hwang, J.~D.; Lee, L.; Da, J.; Zellers, R.;
  Mankoff, R.; and Choi, Y. 2023.
\newblock Do Androids Laugh at Electric Sheep? Humor ``Understanding''
  Benchmarks from {The New Yorker} Caption Contest.
\newblock In \emph{Proc. ACL}, 688--714.

\bibitem[{Hu et~al.(2025)Hu, Dai, Jiang, and Zhou}]{hu2025ablrefl}
Hu, W.-C.; Dai, W.-Z.; Jiang, Y.; and Zhou, Z.-H. 2025.
\newblock Efficient Rectification of Neuro-Symbolic Reasoning Inconsistencies
  by Abductive Reflection.
\newblock In \emph{Proc. AAAI}, 17333--17341.

\bibitem[{Jiao, Huang, and Zhang(2026)}]{jiao2026ember}
Jiao, C.; Huang, S.; and Zhang, Y. 2026.
\newblock Embracing Positional Bias in Multiple-Choice Question Answering via
  Permutation Equivariant Neural Networks.
\newblock In \emph{Proc. AAAI}, 31374--31382.

\bibitem[{Jin et~al.(2019)Jin, Dhingra, Liu, Cohen, and Lu}]{jin2019pubmedqa}
Jin, Q.; Dhingra, B.; Liu, Z.; Cohen, W.~W.; and Lu, X. 2019.
\newblock {PubMedQA}: A Dataset for Biomedical Research Question Answering.
\newblock In \emph{Proc. EMNLP}, 2567--2577.

\bibitem[{Kahneman(2011)}]{kahneman2011thinking}
Kahneman, D. 2011.
\newblock \emph{Thinking, Fast and Slow}.
\newblock Farrar, Straus and Giroux.

\bibitem[{Kwon et~al.(2023)Kwon, Li, Zhuang, Sheng, Zheng, Yu, Gonzalez, Zhang,
  and Stoica}]{kwon2023vllm}
Kwon, W.; Li, Z.; Zhuang, S.; Sheng, Y.; Zheng, L.; Yu, C.~H.; Gonzalez, J.~E.;
  Zhang, H.; and Stoica, I. 2023.
\newblock Efficient Memory Management for Large Language Model Serving with
  {PagedAttention}.
\newblock In \emph{Proc. SOSP}, 611--626.

\bibitem[{Lambert et~al.(2025)Lambert, Morrison, Pyatkin, Huang, Ivison,
  Brahman, Miranda, Liu, Dziri, Lyu et~al.}]{lambert2024tulu3}
Lambert, N.; Morrison, J.; Pyatkin, V.; Huang, S.; Ivison, H.; Brahman, F.;
  Miranda, L. J.~V.; Liu, A.; Dziri, N.; Lyu, X.; et~al. 2025.
\newblock {T{\"u}lu 3}: Pushing Frontiers in Open Language Model Post-Training.
\newblock In \emph{Proc. COLM}.

\bibitem[{Leviathan, Kalman, and Matias(2023)}]{leviathan2023speculative}
Leviathan, Y.; Kalman, M.; and Matias, Y. 2023.
\newblock Fast Inference from Transformers via Speculative Decoding.
\newblock In \emph{Proc. ICML}, 19274--19286.

\bibitem[{Liu et~al.(2026)Liu, Zhang, Luo, Chen, Liu, and
  He}]{liu2026decision1}
Liu, X.; Zhang, H.; Luo, A.; Chen, H.; Liu, X.; and He, B. 2026.
\newblock Decision 1.0: Towards Open Decision Foundation Models.
\newblock \emph{arXiv preprint arXiv:2611.18629}.

\bibitem[{McIlroy-Young et~al.(2024)McIlroy-Young, Brown, Olson, Zhang, and
  Dwork}]{mcilroy2024sbp}
McIlroy-Young, R.; Brown, K.; Olson, C.; Zhang, L.; and Dwork, C. 2024.
\newblock Order-Independence Without Fine Tuning.
\newblock In \emph{Proc. NeurIPS}, 72818--72839.

\bibitem[{Merrill and Sabharwal(2023)}]{merrill2023parallelism}
Merrill, W.; and Sabharwal, A. 2023.
\newblock The Parallelism Tradeoff: Limitations of Log-Precision Transformers.
\newblock \emph{Trans. Assoc. Comput. Linguist.}, 11: 531--545.

\bibitem[{Ong et~al.(2025)Ong, Almahairi, Wu, Chiang, Wu, Gonzalez, Kadous, and
  Stoica}]{ong2024routellm}
Ong, I.; Almahairi, A.; Wu, V.; Chiang, W.-L.; Wu, T.; Gonzalez, J.~E.; Kadous,
  M.~W.; and Stoica, I. 2025.
\newblock {RouteLLM}: Learning to Route {LLMs} from Preference Data.
\newblock In \emph{Proc. ICLR}, 34433--34448.

\bibitem[{Ovadia et~al.(2019)Ovadia, Fertig, Ren, Nado, Sculley, Nowozin,
  Dillon, Lakshminarayanan, and Snoek}]{ovadia2019trust}
Ovadia, Y.; Fertig, E.; Ren, J.; Nado, Z.; Sculley, D.; Nowozin, S.; Dillon,
  J.~V.; Lakshminarayanan, B.; and Snoek, J. 2019.
\newblock Can You Trust Your Model's Uncertainty? Evaluating Predictive
  Uncertainty Under Dataset Shift.
\newblock In \emph{Proc. NeurIPS}, 13991--14002.

\bibitem[{Qiu et~al.(2026)Qiu, Wang, Li, Li, Xu, Wang, Zhang, Men, Mao, Zhang
  et~al.}]{qiu2026qwen38next}
Qiu, Z.; Wang, Z.; Li, X.; Li, Y.; Xu, Y.; Wang, Y.; Zhang, H.; Men, R.; Mao,
  B.; Zhang, C.; et~al. 2026.
\newblock On the Design of {Qwen3.8-Next} Architecture: Evaluation, Efficiency,
  and Training Stability.
\newblock \emph{arXiv preprint arXiv:2608.30320}.

\bibitem[{Reddy et~al.(2022)Reddy, M{\`a}rquez, Valero, Rao, Zaragoza,
  Bandyopadhyay, Biswas, Xing, and Subbian}]{reddy2022esci}
Reddy, C.~K.; M{\`a}rquez, L.; Valero, F.; Rao, N.; Zaragoza, H.;
  Bandyopadhyay, S.; Biswas, A.; Xing, A.; and Subbian, K. 2022.
\newblock Shopping Queries Dataset: A Large-Scale {ESCI} Benchmark for
  Improving Product Search.
\newblock \emph{arXiv preprint arXiv:2206.06588}.

\bibitem[{Ross, Mahabaleshwarkar, and Suhara(2025)}]{ross2025when2call}
Ross, H.; Mahabaleshwarkar, A.~S.; and Suhara, Y. 2025.
\newblock {When2Call}: When (not) to Call Tools.
\newblock In \emph{Proc. NAACL}, 3391--3409.

\bibitem[{Saha et~al.(2025)Saha, Prasad, Chen, Hase, Stengel-Eskin, and
  Bansal}]{saha2024system1x}
Saha, S.; Prasad, A.; Chen, J. C.-Y.; Hase, P.; Stengel-Eskin, E.; and Bansal,
  M. 2025.
\newblock {System-1.x}: Learning to Balance Fast and Slow Planning with
  Language Models.
\newblock In \emph{Proc. ICLR}, 38652--38679.

\bibitem[{Sakshi et~al.(2025)Sakshi, Tyagi, Kumar, Seth, Selvakumar, Nieto,
  Duraiswami, Ghosh, and Manocha}]{sakshi2024mmau}
Sakshi, S.; Tyagi, U.; Kumar, S.; Seth, A.; Selvakumar, R.; Nieto, O.;
  Duraiswami, R.; Ghosh, S.; and Manocha, D. 2025.
\newblock {MMAU}: A Massive Multi-Task Audio Understanding and Reasoning
  Benchmark.
\newblock In \emph{Proc. ICLR}, 84929--84964.

\bibitem[{Schuster et~al.(2022)Schuster, Fisch, Gupta, Dehghani, Bahri, Tran,
  Tay, and Metzler}]{schuster2022calm}
Schuster, T.; Fisch, A.; Gupta, J.; Dehghani, M.; Bahri, D.; Tran, V.~Q.; Tay,
  Y.; and Metzler, D. 2022.
\newblock Confident Adaptive Language Modeling.
\newblock In \emph{Proc. NeurIPS}, 17456--17472.

\bibitem[{Shao et~al.(2024)Shao, Wang, Zhu, Xu, Song, Bi, Zhang, Zhang, Li, Wu,
  and Guo}]{shao2024deepseekmath}
Shao, Z.; Wang, P.; Zhu, Q.; Xu, R.; Song, J.; Bi, X.; Zhang, H.; Zhang, M.;
  Li, Y.~K.; Wu, Y.; and Guo, D. 2024.
\newblock {DeepSeekMath}: Pushing the Limits of Mathematical Reasoning in Open
  Language Models.
\newblock \emph{arXiv preprint arXiv:2402.03300}.

\bibitem[{Singh et~al.(2025)Singh, Romanou, Fourrier, Adelani, Ngui,
  Vila-Suero, Limkonchotiwat, Marchisio, Leong, Susanto, Ng, Longpre, Ruder,
  Ko, Bosselut, Oh, Martins, Choshen, Ippolito, Ferrante, Fadaee, Ermis, and
  Hooker}]{singh2025globalmmlu}
Singh, S.; Romanou, A.; Fourrier, C.; Adelani, D.~I.; Ngui, J.~G.; Vila-Suero,
  D.; Limkonchotiwat, P.; Marchisio, K.; Leong, W.~Q.; Susanto, Y.; Ng, R.;
  Longpre, S.; Ruder, S.; Ko, W.-Y.; Bosselut, A.; Oh, A.; Martins, A. F.~T.;
  Choshen, L.; Ippolito, D.; Ferrante, E.; Fadaee, M.; Ermis, B.; and Hooker,
  S. 2025.
\newblock Global {MMLU}: Understanding and Addressing Cultural and Linguistic
  Biases in Multilingual Evaluation.
\newblock In \emph{Proc. ACL}, 18761--18799.

\bibitem[{Snell et~al.(2025)Snell, Lee, Xu, and Kumar}]{snell2024scaling}
Snell, C.; Lee, J.; Xu, K.; and Kumar, A. 2025.
\newblock Scaling {LLM} Test-Time Compute Optimally Can be More Effective than
  Scaling Parameters for Reasoning.
\newblock In \emph{Proc. ICLR}, 10131--10165.

\bibitem[{Wang et~al.(2023)Wang, Wei, Schuurmans, Le, Chi, Narang, Chowdhery,
  and Zhou}]{wang2023selfconsistency}
Wang, X.; Wei, J.; Schuurmans, D.; Le, Q.~V.; Chi, E.~H.; Narang, S.;
  Chowdhery, A.; and Zhou, D. 2023.
\newblock Self-Consistency Improves Chain of Thought Reasoning in Language
  Models.
\newblock In \emph{Proc. ICLR}.

\bibitem[{Wei et~al.(2022)Wei, Wang, Schuurmans, Bosma, Ichter, Xia, Chi, Le,
  and Zhou}]{wei2022cot}
Wei, J.; Wang, X.; Schuurmans, D.; Bosma, M.; Ichter, B.; Xia, F.; Chi, E.~H.;
  Le, Q.~V.; and Zhou, D. 2022.
\newblock Chain-of-Thought Prompting Elicits Reasoning in Large Language
  Models.
\newblock In \emph{Proc. NeurIPS}, 24824--24837.

\bibitem[{Wolf et~al.(2020)Wolf, Debut, Sanh, Chaumond, Delangue, Moi, Cistac,
  Rault, Louf, Funtowicz, Davison, Shleifer, von Platen, Ma, Jernite, Plu, Xu,
  Le~Scao, Gugger, Drame, Lhoest, and Rush}]{wolf2020transformers}
Wolf, T.; Debut, L.; Sanh, V.; Chaumond, J.; Delangue, C.; Moi, A.; Cistac, P.;
  Rault, T.; Louf, R.; Funtowicz, M.; Davison, J.; Shleifer, S.; von Platen,
  P.; Ma, C.; Jernite, Y.; Plu, J.; Xu, C.; Le~Scao, T.; Gugger, S.; Drame, M.;
  Lhoest, Q.; and Rush, A.~M. 2020.
\newblock Transformers: State-of-the-Art Natural Language Processing.
\newblock In \emph{Proc. EMNLP: System Demonstrations}, 38--45.

\bibitem[{Wu and Lim(2026)}]{wu2026reflex}
Wu, T.; and Lim, W. Y.~B. 2026.
\newblock {REFLEX} with {Jev} for Efficient Selective Control in {LLM} Agents.
\newblock \emph{arXiv preprint arXiv:2609.26532}.

\bibitem[{Xiao et~al.(2026)Xiao, Xu, Yuan, Wang, Li, and Liu}]{xiao2026scale}
Xiao, Y.; Xu, C.; Yuan, R.; Wang, J.; Li, W.; and Liu, P. 2026.
\newblock {SCALE}: Selective Resource Allocation for Overcoming Performance
  Bottlenecks in Mathematical Test-Time Scaling.
\newblock In \emph{Proc. AAAI}, 34034--34042.

\bibitem[{Yang et~al.(2025{\natexlab{a}})Yang, Li, Yang, Zhang, Hui, Zheng, Yu,
  Gao, Huang, Lv et~al.}]{yang2025qwen3}
Yang, A.; Li, A.; Yang, B.; Zhang, B.; Hui, B.; Zheng, B.; Yu, B.; Gao, C.;
  Huang, C.; Lv, C.; et~al. 2025{\natexlab{a}}.
\newblock {Qwen3} Technical Report.
\newblock \emph{arXiv preprint arXiv:2505.09388}.

\bibitem[{Yang et~al.(2026)Yang, Si, Tian, Liu, Yao, Qin, Lin, Wang, and
  Wang}]{yang2026dami}
Yang, C.; Si, Q.; Tian, C.; Liu, X.; Yao, D.; Qin, C.; Lin, Z.; Wang, W.; and
  Wang, J. 2026.
\newblock System 1\&2 Synergy via Dynamic Model Interpolation.
\newblock In \emph{Proc. IJCAI}, 6000--6008.

\bibitem[{Yang, Kautz, and Hatamizadeh(2025)}]{yang2025gateddeltanet}
Yang, S.; Kautz, J.; and Hatamizadeh, A. 2025.
\newblock Gated Delta Networks: Improving {Mamba2} with Delta Rule.
\newblock In \emph{Proc. ICLR}, 29687--29707.

\bibitem[{Yang et~al.(2025{\natexlab{b}})Yang, Ma, Lin, and Wei}]{yang2025tops}
Yang, W.; Ma, S.; Lin, Y.; and Wei, F. 2025{\natexlab{b}}.
\newblock Towards Thinking-Optimal Scaling of Test-Time Compute for {LLM}
  Reasoning.
\newblock In \emph{Proc. NeurIPS}, 48704--48730.

\bibitem[{Zhao et~al.(2021)Zhao, Wallace, Feng, Klein, and
  Singh}]{zhao2021calibrate}
Zhao, T.~Z.; Wallace, E.; Feng, S.; Klein, D.; and Singh, S. 2021.
\newblock Calibrate Before Use: Improving Few-Shot Performance of Language
  Models.
\newblock In \emph{Proc. ICML}, 12697--12706.

\bibitem[{Zheng et~al.(2024)Zheng, Zhou, Meng, Zhou, and
  Huang}]{zheng2024pride}
Zheng, C.; Zhou, H.; Meng, F.; Zhou, J.; and Huang, M. 2024.
\newblock Large Language Models Are Not Robust Multiple Choice Selectors.
\newblock In \emph{Proc. ICLR}.

\bibitem[{Zheng et~al.(2026)Zheng, Yuan, Yao, Gu, Zheng, and
  He}]{zheng2026pagrpo}
Zheng, J.; Yuan, J.; Yao, J.; Gu, C.; Zheng, P.; and He, G. 2026.
\newblock Mitigating Selection Bias in Large Language Models via
  Permutation-Aware {GRPO}.
\newblock In \emph{Proc. ACL}, 35125--35143.

\bibitem[{Zhou et~al.(2026)Zhou, Ling, Chen, Wang, Fan, and
  Wang}]{zhou2026overthinking}
Zhou, S.; Ling, R.; Chen, J.; Wang, X.; Fan, T.; and Wang, H. 2026.
\newblock When More Thinking Hurts: Overthinking in {LLM} Test-Time Compute
  Scaling.
\newblock In \emph{Findings of ACL}, 23967--23977.

\bibitem[{Zhuang et~al.(2026)Zhuang, Su, Pan, Wong, Chen, and
  Kuo}]{zhuang2026startlux}
Zhuang, X.; Su, Z.; Pan, J.; Wong, N.; Chen, G.; and Kuo, C.-W. 2026.
\newblock {StartLux-Decision}: {AutoThink} Decision Models with Reinforcement
  Learning for Calibrated Deliberative Decisions.
\newblock \emph{arXiv preprint arXiv:2610.17309}.

\end{thebibliography}

\ifwithappendix
\clearpage
\appendix
\raggedbottom
\makeatletter
\setlength{\@fptop}{0pt}
\setlength{\@dblfptop}{0pt}
\makeatother
\setcounter{table}{0}
\renewcommand{\thetable}{A\arabic{table}}
\setcounter{figure}{0}
\renewcommand{\thefigure}{A\arabic{figure}}

\section{Evaluation Reproducibility and Training Ablations}
\label{sec:app_repro}

\paragraph{Evaluation Backends and Cross-Backend Consistency.}
The nine primary benchmarks, JevBench-231 evaluations, diagnostic suites, and local 831-item Commercial Suite evaluations run in BF16 configured at 262,144-token context capacity (64K--256K stress tests remain open) on RTX PRO 6000 Blackwell GPUs with greedy decoding, repetition penalty $1.15$, and single-token routing ($\Delta z \le e$, $p_{\max}^{\mathrm{S1}} < 0.8$); Decision Index~0.3 runs single-stream in BF16 via Hugging Face Transformers on an NVIDIA RTX~4080. Primary benchmarks and diagnostics use unified decision prompts, whereas the Commercial Suite evaluates three prompt formats (conversational chat, structured reasoning, and compact decision) per Tractus mode and selects the best format per task. Sequential single-stream inference shares the live prefill KV cache across both modes and resolves escalated answers at the terminal post-trace assistant slot by reading the $K$ reserved option-token logits and matching explicit answer keywords (\texttt{Final Answer}, \texttt{Answer}, \texttt{Choice}, \texttt{Option}, or \texttt{Select} followed by colon/whitespace and optional parenthesized option key, searched after \texttt{</think>} when present) to override $y^{\mathrm{S2}}$ while evaluating $p^{\mathrm{S2}}_{y^{\mathrm{S2}}}$ in both the safety ratchet and ECE calculation. For throughput, our BF16 evaluation harness separates backends: System~1 prefill runs in Transformers, single-stream diagnostics and 8K runs decode System~2 in Transformers with repetition-cycle halting, and high-volume splits plus the Commercial Suite send escalated queries to continuous-batching vLLM (without cross-engine KV reuse) before terminal-slot readout and answer extraction. Independent BF16 runs in our harness agree within $0.08$\,pp weighted accuracy. Padding and numerical differences up to $0.125$\,nats between batched vLLM 4K runs (Table~\ref{tab:app_4k_fig4}a, Figure~\ref{fig:scaling_panorama}a) and single-stream Transformers ablations (Table~\ref{tab:unified_panorama} JevBench-231, Table~\ref{tab:app_4k_fig4}b, Figure~\ref{fig:scaling_panorama}b) explain single-item System~1 boundary flips ($161$ vs.\ $160$ at 2B; $182$ vs.\ $183$ at 9B; $176/231$ unchanged at 4B) and Dual-Mode shifts ($150$ at 4K vs.\ $151$ at 8K for 2B; $201$ vs.\ $200$ at 4K for 4B). They also explain one-item System~2 escalation differences between the 8K (Table~\ref{tab:ablation}b) and 4K (Table~\ref{tab:app_4k_fig4}b) runs ($29.0\%$ vs.\ $29.4\%$ at 9B; $56.3\%$ vs.\ $55.8\%$ at 4B) and a substantial $1.84$\,pp 4B System~1 ECE discrepancy between Table~\ref{tab:ablation}a ($8.29\%$) and Table~\ref{tab:app_4k_fig4}b ($10.13\%$)---exceeding Table~\ref{tab:ablation}a's $0.31$\,pp thermostat step---while leaving Brier score unchanged at $0.323$. Comparing uniform 4K runs against Table~\ref{tab:unified_panorama}'s 8K JevBench totals contrasts the 4K and 8K generation caps alongside these minor backend shifts.

\paragraph{Thermostasis Inference Pseudocode.}
Algorithm~\ref{alg:thermostasis} details the dual-mode Thermostasis inference procedure from Section~\ref{sec:thermostasis}, denoting the margin threshold $\tau_z=e$, confidence guard $\tau_p=0.8$, ratchet floor $\tau_r=0.4$, and token length $L_{\mathrm{gen}}(\mathbf{r})$ of reasoning trace $\mathbf{r}$.

\begin{algorithm}[t]
\caption{Thermostasis Dual-Mode Inference}
\label{alg:thermostasis}
\begin{algorithmic}[1]
\REQUIRE Input context, $K$-candidate menu with $2 \le K \le 255$, static prior $\mathbf{z}_0$, mode $\in \{\mathrm{Joint}, \mathrm{GO}\}$, constants $C{=}30, e^{-1}, \tau_z{=}e, \tau_p{=}0.8, \tau_r{=}0.4$
\STATE Extract candidate logits $\mathbf{z}_{\mathrm{raw}}$ at prefill
\STATE Debias and softcap $\bar{\mathbf{z}} \leftarrow C\tanh((\mathbf{z}_{\mathrm{raw}} - \mathbf{z}_0)/C)$; zero-sum center $\tilde{\mathbf{z}}$ via Eq.~\eqref{eq:centering}
\STATE Sort $\tilde{\mathbf{z}}$ descending; compute $\Delta z \leftarrow \tilde{z}_{(1)} - \tilde{z}_{(2)}$ and $S_{\mathrm{r}} \leftarrow \log(1 + \sum_{j=3}^K e^{\tilde{z}_{(j)}-\tilde{z}_{(2)}})$ ($0$ if $K=2$)
\STATE Compute $T(\Delta z, S_{\mathrm{r}})$ via Eq.~\eqref{eq:ethermo}
\STATE Set $\mathbf{p}^{\mathrm{S1}} \leftarrow \mathrm{softmax}(\tilde{\mathbf{z}}/T)$, $y^{\mathrm{S1}} \leftarrow \arg\max_j \tilde{z}_j$, $p_{\max}^{\mathrm{S1}} \leftarrow p^{\mathrm{S1}}_{y^{\mathrm{S1}}}$
\IF{\rule[-0.4ex]{0pt}{2.2ex}$\Delta z > \tau_z$ \textbf{or} (mode is $\mathrm{Joint}$ \textbf{and} $p_{\max}^{\mathrm{S1}} \ge \tau_p$)}
    \STATE \textbf{return} $(y^{\mathrm{S1}}, \mathbf{p}^{\mathrm{S1}}, 0)$ \hfill $\triangleright$ single-pass System~1 exit (no decoding)
\ENDIF
\STATE Decode trace $\mathbf{r}$ of $L_{\mathrm{gen}}(\mathbf{r})$ tokens from shared prefix; read $\mathbf{z}_{\mathrm{raw}}^{\mathrm{S2}}$
\STATE Debias, softcap, and calibrate $\mathbf{z}_{\mathrm{raw}}^{\mathrm{S2}}$ via Eqs.~\eqref{eq:centering}--\eqref{eq:ethermo} for $(y^{\mathrm{S2}}, \mathbf{p}^{\mathrm{S2}})$; override $y^{\mathrm{S2}}$ if an explicit answer delimiter is matched
\IF{\rule{0pt}{2.2ex}$p_{y^{\mathrm{S2}}}^{\mathrm{S2}} \ge \max(\tau_r, 1/K)$}
    \STATE \textbf{return} $(y^{\mathrm{S2}}, \mathbf{p}^{\mathrm{S2}}, L_{\mathrm{gen}}(\mathbf{r}))$
\ELSE
    \STATE \textbf{return} $(y^{\mathrm{S1}}, \mathbf{p}^{\mathrm{S1}}, L_{\mathrm{gen}}(\mathbf{r}))$ \hfill $\triangleright$ safety ratchet
\ENDIF
\end{algorithmic}
\end{algorithm}

\paragraph{Inference Settings, Binary Priors, and Training Ablations.}
Main Tractus inference evaluations reproduce from the released BF16 checkpoints within the cross-backend tolerance above (262,144-token capacity, greedy decoding, repetition penalty $1.15$, $2 \le K \le 255$ candidate readout, five fixed Thermostasis constants $C{=}30, e^{-1}, \tau_z{=}e, \tau_p{=}0.8, \tau_r{=}0.4$, no fitted temperatures by default). Before Eq.~\eqref{eq:centering}, logits subtract optional antisymmetric prior $\mathbf{z}_0 = (+\Delta z_0/2, -\Delta z_0/2)$ and undergo softcapping $\bar{\mathbf{z}} = C\tanh((\mathbf{z}_{\mathrm{raw}} - \mathbf{z}_0)/C)$. Dev-estimated static priors (sensitive to base-rate shift) apply only at 9B on two Decision Index~0.3 catalogs: $\Delta z_0{=}+3.2$\,nats on RAGTruth (Catalog~59, positive: \textit{``All content of the response is supported by the context in the prompt.''}; negative: \textit{``The response contains content that is not supported by the context in the prompt.''}) and $\Delta z_0{=}-0.75$\,nats on PhishNChips (Catalog~56, \textit{phishing}/\textit{legitimate}, adding $+0.375$\,nats to \textit{phishing}); $\mathbf{z}_0{=}\mathbf{0}$ across 0.8B--4B and on the primary, JevBench, OOD, and remaining Decision Index splits. Matching Section~\ref{sec:training}, post-training from Qwen3.5-Base differs by scale: 0.8B/2B complete Stage-1 distillation/margin alignment and Stage-2 CoT with ${\sim}25\%$ replay and tiered LRs (zero KL); 4B and 9B add Stage-3 interactive RLVR and binary ATP-RLVR ($\epsilon{=}0.2, \beta_{\mathrm{KL}}{=}0.04, \lambda_{\mathrm{m}}{=}0.1$, rewards $+1.0/{-}2.0/{-}2.5/{-}1.5$, top blocks/head only), with 9B also adding a brief Stage-1 refinement. Proprietary Stage-1/2 schedules, corpus counts, and decontamination counts are omitted from the public release. Table~\ref{tab:app_train_ablations} reports training ablations under pre-standardization prompts (explaining Table~\ref{tab:unified_panorama} baseline shifts: 2B $75.76\%$ vs.\ $69.26\%$; 9B $82.25\%$ vs.\ $79.22\%$): tiered Stage-2 LRs retain $175/231$ JevBench vs.\ $167$ uniform on the 128-item 2B holdout, while 9B ATP-RLVR cuts phishing bias from $-3.66$ to $-0.75$\,nats without observed Web-Trajectory ($11/12$, $0$ premature exits) or General NLU ($640/720$, $419/498$) regressions.

\begin{table*}[t]
\centering
\small
\begin{minipage}[t]{0.48\textwidth}
\vspace{0pt}
\centering
\renewcommand{\arraystretch}{0.86}
\setlength{\tabcolsep}{2.2pt}
\begin{tabular*}{\linewidth}{@{\extracolsep{\fill}} lccc @{}}
\toprule
\multicolumn{4}{@{}l}{\textbf{(a) 2B Stage-2 Tiered-LR Study (128-Item Web Holdout)}} \\
\midrule
\textbf{Variant} & \textbf{Web Holdout} & \textbf{JevBench-231} & \textbf{Delta vs.\ Base} \\
\midrule
2B Base & -- & 75.76\% (175) & Baseline \\
2B Unif. & 85.16\% & 167 ($-3.46$\,pp) & $-8$ items \\
2B Tiered & 85.16\% & 175 (0.00\,pp) & $0$ items \\
\bottomrule
\end{tabular*}
\end{minipage}\hfill
\begin{minipage}[t]{0.49\textwidth}
\vspace{0pt}
\centering
\renewcommand{\arraystretch}{0.86}
\setlength{\tabcolsep}{2.2pt}
\begin{tabular*}{\linewidth}{@{\extracolsep{\fill}} lcccc @{}}
\toprule
\multicolumn{5}{@{}l}{\textbf{(b) 9B Stage-3 ATP-RLVR (37-Catalog DI \& $\Delta z_0$)}} \\
\midrule
\textbf{Stage} & \textbf{NLU} & \textbf{JB-231} & \textbf{DI (37)} & \textbf{Phish.\ $\Delta z_0$} \\
\midrule
9B S2 & 86.86\% & 82.25\% (190) & 48.51 & $-3.66$\,nats \\
9B RLVR & 86.95\% & 82.25\% (190) & 48.46 & $-0.75$\,nats \\
\bottomrule
\end{tabular*}
\end{minipage}
\caption{Training ablations for Stage-2 retention and Stage-3 debiasing (evaluated under earlier pre-standardization prompts, explaining JevBench baseline differences vs.\ Table~\ref{tab:unified_panorama}'s unified-prompt 2B $69.26\%$ and 9B $79.22\%$): (a)~2B Stage-2 tiered-LR study on the 128-item web holdout, and (b)~9B Stage-3 retention and phishing prior offset $\Delta z_0$ (NLU: General NLU Suite, 1,218 items; JB-231: JevBench-231; DI~(37): 37-catalog Decision Index~0.3 subset).}
\label{tab:app_train_ablations}
\end{table*}

\section{Full Decision Index 0.3 Results}
\label{sec:app_di}

Table~\ref{tab:app_decision_index} reports the full Decision Index~0.3 comparison against 18 public baselines alongside Breadth scores weighted toward unseen domains. Decision Index~0.3 weights five official domains---Tools ($18.3\%$), Language ($25.8\%$), Retrieval ($20.0\%$), Knowledge ($25.8\%$), and Arts \& Human Taste ($10.0\%$; summing to $99.9\%$ by one-decimal rounding)---and \tractusdec\ leads the compared baselines in Balanced Skill across the four evaluated tiers, whereas specialized baselines such as Cloudflare clef-flash post higher Balanced Raw accuracy at lower Balanced Skill.

\begin{table*}[t]
\centering
\small
\renewcommand{\arraystretch}{0.80}
\setlength{\tabcolsep}{3.6pt}
\begin{tabular*}{\textwidth}{@{\extracolsep{\fill}} llcccc @{}}
\toprule
\multicolumn{6}{@{}l}{\textbf{(a) Full 22-Model Decision Index 0.3 Leaderboard (2,172 Items, 43 Active Catalogs)}} \\
\midrule
Tier & Model & Size & Bal.\ Skill $\uparrow$ & Bal.\ Raw $\uparrow$ & Breadth $\uparrow$ \\
\midrule
\multirow{7}{*}{9B}
& \textbf{Tractus-decision-9B (Ours)} & 9.24B & \textbf{46.03} & 59.32\% & 44.71 \\
& vLLM-SR Decision 2.0 Lux 9B & 9.65B & 45.97 & 59.93\% & 45.87 \\
& JevK5 9B v0.3.3 & 9.65B & 44.89 & 60.38\% & \textbf{46.33} \\
& Cloudflare clef-flash & 9.65B & 44.43 & \textbf{66.53\%} & -- \\
& JPT-9B & 9.65B & 44.08 & 59.66\% & 45.16 \\
& JEV-9B & 9.65B & 42.15 & 57.47\% & 42.23 \\
& Sieve 9B & 9.65B & 41.73 & 57.12\% & 40.31 \\
\midrule
\multirow{6}{*}{4B}
& \textbf{Tractus-decision-4B (Ours)} & 4.21B & \textbf{41.31} & 55.56\% & 40.00 \\
& vLLM-SR Decision 2.0 Nox 4B & 4.21B & 40.75 & \textbf{58.08\%} & -- \\
& Hopper (G) 1.2 (4B) & 4.66B & 40.17 & 56.00\% & -- \\
& JPT-4B & 4.66B & 39.56 & 56.74\% & -- \\
& Intern-Decision-4B & 4.66B & 38.69 & 53.59\% & -- \\
& Decider 4B & 4.66B & 36.30 & 56.40\% & -- \\
\midrule
\multirow{6}{*}{2B}
& \textbf{Tractus-decision-2B (Ours)} & 1.88B & \textbf{29.27} & 46.83\% & 26.73 \\
& this-that 1.2 (2B) & 1.88B & 24.85 & 45.76\% & -- \\
& Decider 2B & 2.27B & 23.65 & \textbf{47.01\%} & -- \\
& vLLM-SR Decision 2.0 Sol 2B & 2.27B & 23.64 & 46.89\% & -- \\
& Sieve 2B & 2.27B & 21.90 & 40.70\% & -- \\
& Intern-Decision-2B & 2.27B & 20.93 & 37.89\% & -- \\
\midrule
\multirow{3}{*}{Sub-2B}
& \textbf{Tractus-decision-0.8B (Ours)} & 0.82B & \textbf{18.07} & 37.66\% & 17.10 \\
& Bosun v3.1 1.7B & 1.72B & 16.54 & \textbf{41.06\%} & -- \\
& JPT-0.8B & 0.87B & 15.46 & 39.36\% & -- \\
\bottomrule
\end{tabular*}

\vspace{4pt}
\setlength{\tabcolsep}{3.0pt}
\begin{tabular*}{\textwidth}{@{\extracolsep{\fill}} lccccccc @{}}
\toprule
\multicolumn{8}{@{}l}{\textbf{(b) Tractus Per-Domain Skill (Raw Acc.\%) and Weighted Balanced Skill Breakdown}} \\
\midrule
Model & Tools (18.3\%) & Lang.\ (25.8\%) & Retr.\ (20.0\%) & Know.\ (25.8\%) & Taste (10.0\%) & Breadth & Bal.\ Skill \\
\midrule
\textbf{Tractus-decision-9B (Ours)} & \thdr{\textbf{58.15}\\(62.0\%)} & \thdr{\textbf{60.40}\\(73.2\%)} & \thdr{37.44\\(54.8\%)} & \thdr{\textbf{31.71}\\(47.3\%)} & \thdr{\textbf{40.95}\\(58.5\%)} & \textbf{44.71} & \textbf{46.03} \\
\addlinespace[2pt]
\textbf{Tractus-decision-4B (Ours)} & \thdr{53.63\\(58.8\%)} & \thdr{53.19\\(67.7\%)} & \thdr{\textbf{37.97}\\(53.2\%)} & \thdr{26.06\\(44.0\%)} & \thdr{34.19\\(52.9\%)} & 40.00 & 41.31 \\
\addlinespace[2pt]
\textbf{Tractus-decision-2B (Ours)} & \thdr{43.24\\(49.0\%)} & \thdr{41.16\\(59.3\%)} & \thdr{31.79\\(49.7\%)} & \thdr{11.52\\(33.1\%)} & \thdr{13.79\\(40.5\%)} & 26.73 & 29.27 \\
\addlinespace[2pt]
\textbf{Tractus-decision-0.8B (Ours)} & \thdr{23.84\\(32.6\%)} & \thdr{27.65\\(49.0\%)} & \thdr{15.26\\(33.3\%)} & \thdr{10.45\\(33.7\%)} & \thdr{8.10\\(36.5\%)} & 17.10 & 18.07 \\
\bottomrule
\end{tabular*}
\caption{Full Decision Index~0.3 comparison against 18 public baselines across 43 catalogs covering $1{,}932$ indexed plus $240$ unindexed reference items; dashes indicate unpublished baseline Breadth scores.}
\label{tab:app_decision_index}
\end{table*}

\section{Decontamination and Extended Results}
\label{sec:app_corpus}

\subsection{Decontamination Protocol and Domain Coverage}
Post-training data cover four operational decision domains: fact verification and long-document compliance auditing; intent routing, retrieval relevance, and safety screening; tool selection, multi-step reasoning, and dialogue; and multimodal document, chart, GUI, and visual scene decision tasks, paired with counterfactual twin variants. To mitigate data leakage, training splits are filtered against our evaluation benchmarks via Disjoint-Set Union clustering over semantic twin identifiers alongside normalized 13-gram overlap matching and MinHash-LSH deduplication covering JevBench-231, Decision Index~0.3, and the reported diagnostic suites.

\begin{table*}[t]
\centering
\small
\renewcommand{\arraystretch}{0.75}
\setlength{\tabcolsep}{2.4pt}
\begin{tabular*}{\textwidth}{@{\extracolsep{\fill}} llccccccc @{}}
\toprule
Model & Setting & wAcc $\uparrow$ & wPrim $\uparrow$ & wECE $\downarrow$ & CycFlip $\downarrow$ & \thdr{W2C\\(3,652)} & \thdr{MMLU-L\\(9,200)} & \thdr{ESCI-F1\\(5,000)} \\
\midrule
\multirow{4}{*}{Tractus-9B}
& Raw Logits & 68.09\% & 62.60\% & 13.77\% & 22.31\% & 75.36\% & 67.77\% & 38.83\% \\
& Softcap \& Center ($T{=}1$) & 68.09\% & 62.60\% & \textbf{6.85\%} & 22.31\% & 75.36\% & 67.77\% & 38.83\% \\
& Margin-Only Thermostat & 68.09\% & 62.60\% & 8.21\% & 22.31\% & 75.36\% & 67.77\% & 38.83\% \\
& Tractus-dec. (2D) & 68.09\% & 62.60\% & 8.52\% & 22.31\% & 75.36\% & 67.77\% & 38.83\% \\
\midrule
\multirow{4}{*}{Tractus-4B}
& Raw Logits & 62.66\% & 57.75\% & 10.62\% & 29.17\% & 64.90\% & 61.40\% & 41.87\% \\
& Softcap \& Center ($T{=}1$) & 62.66\% & 57.75\% & \textbf{8.28\%} & 29.17\% & 64.90\% & 61.40\% & 41.87\% \\
& Margin-Only Thermostat & 62.66\% & 57.75\% & 11.41\% & 29.17\% & 64.90\% & 61.40\% & 41.87\% \\
& Tractus-dec. (2D) & 62.66\% & 57.75\% & 10.80\% & 29.17\% & 64.90\% & 61.40\% & 41.87\% \\
\midrule
\multirow{4}{*}{Tractus-2B}
& Raw Logits & 51.87\% & 45.76\% & 6.73\% & 36.42\% & 58.73\% & 44.48\% & 30.76\% \\
& Softcap \& Center ($T{=}1$) & 51.87\% & 45.76\% & \textbf{5.79\%} & 36.42\% & 58.73\% & 44.48\% & 30.76\% \\
& Margin-Only Thermostat & 51.87\% & 45.76\% & 5.91\% & 36.42\% & 58.73\% & 44.48\% & 30.76\% \\
& Tractus-dec. (2D) & 51.87\% & 45.76\% & 6.03\% & 36.42\% & 58.73\% & 44.48\% & 30.76\% \\
\midrule
\multirow{4}{*}{Tractus-0.8B}
& Raw Logits & 43.81\% & 37.67\% & 35.37\% & 42.42\% & 39.43\% & 38.15\% & 26.98\% \\
& Softcap \& Center ($T{=}1$) & 43.81\% & 37.67\% & 28.85\% & 42.42\% & 39.43\% & 38.15\% & 26.98\% \\
& Margin-Only Thermostat & 43.81\% & 37.67\% & \textbf{28.29\%} & 42.42\% & 39.43\% & 38.15\% & 26.98\% \\
& Tractus-dec. (2D) & 43.81\% & 37.67\% & 29.20\% & 42.42\% & 39.43\% & 38.15\% & 26.98\% \\
\bottomrule
\end{tabular*}
\caption{Single-pass calibration across seven full public OOD splits totaling $19{,}887$ items without System~2 ($\mathbf{z}_0{=}\mathbf{0}$, $C{=}30$): When2Call ($3{,}652$), Global-MMLU-Lite ($9{,}200$), Amazon ESCI ($5{,}000$), PubMedQA ($1{,}000$), New Yorker Caption ($528$), Mind2Web ($276$), and JevBench-231 ($231$). Columns report weighted accuracy (wAcc), primary score (wPrim: ESCI Macro-F1 and Top-1 elsewhere), 15-bin wECE, and cyclic option flip rate (CycFlip, invariant under rank-preserving Thermostasis); \textbf{bold} marks lowest wECE per scale.}
\label{tab:app_stage_c}
\end{table*}

\subsection{Single-Pass Calibration and 4K Routing Ablations}
Softcapping at $C{=}30$ ($T{=}1$) preserves top-1 predictions while achieving the lowest tested single-pass OOD calibration error across 2B--9B in Table~\ref{tab:app_stage_c}, with $\mathbf{z}_0{=}\mathbf{0}$ throughout. Because post-softcap zero-sum centering Eq.~\eqref{eq:centering} is shift-invariant and leaves $T{=}1$ softmax probabilities unchanged, bounding extreme logits via $\tanh$ softcapping accounts for the $T{=}1$ OOD wECE reduction over Raw Logits. By contrast, 2D $e$-thermostat cooling sharpens decisive leads for downstream System~2 triage, raising single-pass OOD wECE relative to $T{=}1$ at 4B--9B ($8.28\%{\to}10.80\%$ at 4B; $6.85\%{\to}8.52\%$ at 9B), while rank preservation leaves CycFlip constant at $22.31\%\text{--}42.42\%$.

\begin{table*}[tp]
\centering
\small
\renewcommand{\arraystretch}{0.71}
\setlength{\tabcolsep}{3.0pt}
\begin{tabular*}{\textwidth}{@{\extracolsep{\fill}} lccccc @{}}
\toprule
\multicolumn{6}{@{}l}{\textbf{(a) Uniform 4K Nine-Benchmark Evaluation (9,835 Items/Scale; Data for Figure~\ref{fig:scaling_panorama}a)}} \\
\midrule
\textbf{Scale} & \textbf{4K Macro (S1 $\to$ Dual)} & \textbf{4K Pooled (S1 $\to$ Dual)} & \textbf{Total Esc.\ (\% $\ge$4K)} & \textbf{Parse Fail (Count)} & \textbf{Dual Mean Tok} \\
\midrule
0.8B & 47.84\% $\to$ 47.23\% & 46.22\% $\to$ 45.19\% & 4,846 (79.4\%) & 99.6\% (4,827) & 1,605.8 \\
2B   & 55.77\% $\to$ 54.26\% & 54.92\% $\to$ 53.18\% & 8,265 (51.9\%) & 66.4\% (5,486) & 1,832.9 \\
4B   & 66.06\% $\to$ 72.53\% & 64.57\% $\to$ 72.05\% & 8,554 (4.6\%)  & 13.2\% (1,125) & 970.2   \\
9B   & 71.41\% $\to$ 76.91\% & 70.52\% $\to$ 76.15\% & 4,958 (2.0\%)  & 13.9\% (689)   & 547.5   \\
\bottomrule
\end{tabular*}

\vspace{4pt}
\setlength{\tabcolsep}{2.2pt}
\begin{tabular*}{\textwidth}{@{\extracolsep{\fill}} llcccccc @{}}
\toprule
\multicolumn{8}{@{}l}{\textbf{(b) Matched 4K Routing Variants at 9B and 4B (JevBench-231 and Five-Domain Suite; Data for Figure~\ref{fig:scaling_panorama}b)}} \\
\midrule
\textbf{Scale} & \textbf{Routing Variant} & \textbf{JB (231)} & \textbf{JB ECE} & \textbf{JB Tok\,(Neg)} & \textbf{5-Dom. (550)} & \textbf{5-Dom. wECE} & \textbf{5-Dom. Tok} \\
\midrule
\multirow{6}{*}{\textbf{9B}}
& Tractus-dec. (S1) & 79.22\% (0.0\%) & 7.10\% & 0 (0) & 79.45\% (0.0\%) & \textbf{10.01\%} & 0 \\
& Uncal. Router & 88.31\% (21.2\%) & 8.83\% & 484 (1) & \textbf{80.55\%} (29.8\%) & 17.65\% & 362 \\
& \textbf{Tractus} & 89.61\% (29.4\%) & 7.38\% & \textbf{666 (2)} & 79.27\% (43.3\%) & 19.55\% & \textbf{516} \\
& GO & 92.21\% (42.9\%) & 5.56\% & 874 (4) & 79.45\% (56.7\%) & 19.41\% & 652 \\
& Free-Energy Router & 92.64\% (49.8\%) & 5.25\% & 918 (4) & 79.45\% (72.5\%) & 19.77\% & 765 \\
& Unconstrained CoT & \textbf{93.07\%} (100.0\%) & \textbf{4.20\%} & 1,128 (6) & 79.09\% (100.0\%) & 20.27\% & 919 \\
\midrule
\multirow{6}{*}{\textbf{4B}}
& Tractus-dec. (S1) & 76.19\% (0.0\%) & 10.13\% & 0 (0) & 74.00\% (0.0\%) & 17.49\% & 0 \\
& Uncal. Router & 81.39\% (29.0\%) & 12.95\% & 693 (6) & \textbf{78.55\%} (39.3\%) & \textbf{17.00\%} & 550 \\
& \textbf{Tractus} & 86.58\% (55.8\%) & 8.19\% & \textbf{1,058 (8)} & \textbf{78.55\%} (77.8\%) & 21.48\% & \textbf{886} \\
& GO & 87.45\% (81.8\%) & \textbf{7.93\%} & 1,179 (9) & \textbf{78.55\%} (96.2\%) & 20.45\% & 1,005 \\
& Free-Energy Router & 87.45\% (100.0\%) & 8.40\% & 1,228 (9) & \textbf{78.55\%} (100.0\%) & 20.49\% & 1,019 \\
& Unconstrained CoT & \textbf{87.88\%} (100.0\%) & 8.03\% & 1,228 (11) & 78.18\% (100.0\%) & 20.37\% & 1,019 \\
\bottomrule
\end{tabular*}
\caption{Uniform 4K scaling and routing ablations supporting Figure~\ref{fig:scaling_panorama}. (a)~9,835-item 4K evaluation with Dual-Mode mean generated tokens; batched vLLM frequently hits the 4,096-token cap below 4B, whereas single-stream decoding halts repetitive loops early. (b)~Matched 4K routing variants at 4B--9B, reporting accuracy (System~2 escalation rate) and Tok\,(Neg) as mean tokens (negative flips); \textbf{bold} marks best accuracy/ECE and the calibrated Tractus token cost.}
\label{tab:app_4k_fig4}
\end{table*}

Uniform 4K evaluation across the nine benchmarks indicates frequent sub-4B System~2 parse failures ($99.6\%$ at 0.8B; $66.4\%$ at 2B), whereas 4B and 9B convert escalation into net aggregate gains (Table~\ref{tab:app_4k_fig4}a). Specifically, the uniform 4,096-token evaluation shifts JevBench-231 accuracy from System~1 to Dual-Mode from $72.73\%$ to $70.13\%$ at 0.8B, $69.70\%$ to $64.94\%$ at 2B, $76.19\%$ to $87.01\%$ at 4B, and $78.79\%$ to $89.61\%$ at 9B. Below 4B, batched vLLM frequently hits the 4,096-token generation cap on escalated traces, whereas cycle detection halts repetitive decoding early.

Although Free-Energy Router and GO raise 4K JevBench accuracy (Table~\ref{tab:app_4k_fig4}b), both sharply increase escalation. Free-Energy Router escalates whenever $\Delta F \le e$, omitting both $\tau_p$ and $\Delta z > e$ exits, so it escalates Noisy-Tail items with $>95\%$ S1 accuracy ($95.4\%$ 9B Five-Domain; $100\%$ 4B JB), pushing 4B to $100\%$ S2 on both suites, 9B Five-Domain to $72.5\%$, and worsening untabulated 0.8B failures. GO keeps Noisy-Tail exits but drops $p_{\max}^{\mathrm{S1}} \ge 0.8$. Because 4B logits are smaller, no evaluated 4B query satisfies $\Delta F > e$; items averaging $92.27\%$ S1 accuracy on $\tau_p$-protected Pooled 9-Task ($96.04\%$ 4B Five-Domain) stay in S1 only via $\tau_p$, and dropping $\tau_p$ lifts 4B escalation from $55.8\%$ to $81.8\%$ (JB) and $77.8\%$ to $96.2\%$ (Five-Domain) with zero Five-Domain gain. At 9B, $97.4\%$ of easy queries satisfy $\Delta F > e$, leaving a harder JB subset where GO helps, but GO doubles 9B JB negative flips ($2$ to $4$), lowers Commercial Suite PubMedQA ($70\%$ to $65\%$) and ESCI ($59.23\%$ to $55.75\%$; Table~\ref{tab:commercial_suite}a), degrades untabulated sub-4B Five-Domain accuracy, and pushes 9B Five-Domain escalation past $50\%$ ($43.3\%$ to $56.7\%$), shifting PubMedQA and non-English MMLU to majority decoding for $+1$ item. Keeping $(\tau_z,\tau_p)=(e, 0.8)$ preserves one fixed rule across scales and domains, whereas compute-unconstrained 9B reasoning may relax $\tau_p$.

\subsection{Commercial and Qwen3.5 Suite Breakdown}
Table~\ref{tab:commercial_suite} expands Table~\ref{tab:decision_index}b across 0.8B--9B fixed-order runs of \tractusdec, Dual-Mode \tractus, 9B GO, and Qwen3.5-9B/4B Thinking. External APIs use DataCamp full-split scores (\S\ref{sec:protocol}) and are excluded from local bolding; Tractus uses best-of-3 prompts vs.\ 1-prompt Qwen3.5. ECE uses DataCamp's 10-bin convention (non-comparable to \S\ref{sec:experiments}'s 15-bin); optional 5-fold temperature scaling cuts ECE from $4.99\%$ to $1.07\%$ (4B Dual), $2.80\%$ to $2.69\%$ (9B GO), and $2.74\%$ to $1.97\%$ (2B \tractusdec). On this fixed-order 3-prompt suite at 4B--9B (Table~\ref{tab:commercial_suite}b), \tractusdec\ leads ESCI, Mind2Web, New Yorker, and PubMedQA (unlike Table~\ref{tab:unified_panorama}'s unified-prompt 9B Dual-Mode lead on the first three), while Dual-Mode aids JB, MMLU-Lite, and When2Call. On the $n{=}70$ ($7{\times}10$) sequential BF16 probe, Qwen3.5-9B Thinking averages $2{,}924.6$ tokens ($2{,}924.8$ at 4B; both round to $2{,}925$) at $26.00$/$34.01$/$97.55$\,s med/mean/p95 with $44/831$ full-suite extraction failures. Single-pass \tractusdec-9B runs at $47.4$\,ms/prompt med ($69.2$\,ms mean; $0.14$/$0.18$/$0.29$\,s, 3 prompts); \tractus-9B resolves $479/831$ items without decoding ($352$ escalated) at $0.69$/$7.65$/$27.48$\,s with $0$ failures via logit recovery, while GO escalates $480$ items ($92.21\%$ JB, $7.22$/$10.07$/$28.09$\,s) and drops PubMedQA from $70\%$ to $65\%$.

Five-option probability vectors for \textit{Multi-Rule Approval} in Figure~\ref{fig:qualitative_case}b, ordered as CFO, Exec.\ Comm., Dept.\ Head, Budget Holder, and VP \& Fin.\ Dir., correspond to (0.1590, 0.3542, 0.2330, 0.1269, 0.1269) under 4B System~1, (0.0417, 0.3875, 0.3408, 0.1252, 0.1048) under rejected 2B System~2, and (0.9988, 0.0000, 0.0009, 0.0003, 0.0000) under 4B System~2.

\begin{table*}[tp]
\centering
\small
{\renewcommand{\arraystretch}{0.70}
\setlength{\tabcolsep}{1.5pt}
\begin{tabular*}{\textwidth}{@{\extracolsep{\fill}} lccccccccc @{}}
\toprule
\multicolumn{10}{@{}l}{\textbf{(a) Local 831-Task Suite (Matched BF16 GPUs, Fixed Order) vs.\ External Full-Split API Reference}} \\
\midrule
& \multicolumn{3}{c}{\textbf{Tractus-9B (3-Prompt)}} & \multicolumn{2}{c}{\textbf{Qwen3.5 (1-Prompt)}} & \multicolumn{4}{c}{\textit{External Full-Split APIs (DataCamp)}} \\
\cmidrule(lr){2-4}\cmidrule(lr){5-6}\cmidrule(lr){7-10}
Benchmark or Task & \thdr{Tractus-\\dec.-9B} & \thdr{Tractus-\\9B} & \thdr{Tractus-\\9B (GO)} & \thdr{Qwen3.5\\9B (Th.)} & \thdr{Qwen3.5\\4B (Th.)} & \thdr{Gemini\\Dec.\ 1} & \thdr{Jev\\1.14} & \thdr{GPT-6\\Luna} & \thdr{Flash-\\Lite} \\
\midrule
JevBench (Public, 231) & 80.52\% & 88.74\% & \underline{92.21\%} & \textbf{93.07\%} & 89.18\% & 92.2\% & 90.0\% & 90.9\% & 86.6\% \\
Amazon ESCI (F1, 100) & \underline{63.12\%} & 59.23\% & 55.75\% & \textbf{65.23\%} & 62.69\% & 58.7\% & 56.7\% & 54.7\% & 47.2\% \\
New Yorker Caption (100) & \textbf{78.00\%} & 65.00\% & 64.00\% & 68.00\% & 66.00\% & 79.7\% & 72.0\% & 68.4\% & 73.7\% \\
Mind2Web Multimodal (100) & \textbf{88.00\%} & 85.00\% & 85.00\% & 79.00\% & 78.00\% & 86.2\% & 86.8\% & 81.8\% & 78.0\% \\
PubMedQA (100) & \textbf{76.00\%} & 70.00\% & 65.00\% & 74.00\% & 70.00\% & 79.8\% & 78.2\% & 74.4\% & 77.6\% \\
Global-MMLU-Lite (100) & 67.00\% & 77.00\% & \textbf{79.00\%} & 72.00\% & 71.00\% & 86.0\% & 85.0\% & 79.6\% & 78.8\% \\
When2Call (100) & 75.00\% & \textbf{84.00\%} & \textbf{84.00\%} & \textbf{84.00\%} & 81.00\% & 79.9\% & 82.0\% & 70.4\% & 60.9\% \\
\midrule
Unfitted 10-Bin ECE (7 shared) $\downarrow$ & \textbf{2.45\%} & 5.28\% & 2.80\% & -- & -- & 2.50\% & 4.94\% & 7.21\% & 21.04\% \\
Parse / Extr.\ Failures (of 831) $\downarrow$ & \textbf{0}\textsuperscript{\S} & \textbf{0}\textsuperscript{\S} & \textbf{0}\textsuperscript{\S} & 44 & 65 & -- & -- & -- & -- \\
S2 Esc.\ (831) / Qwen Tok ($n{=}70$) & 0 & 352 & 480 & 2,925 & 2,925 & -- & -- & -- & -- \\
Seq.\ Med.\ Lat ($n{=}70$ Probe) $\downarrow$ & \textbf{0.14\,s}\textsuperscript{\ddag} & 0.69\,s & 7.22\,s & 26.00\,s & 14.36\,s & -- & -- & -- & -- \\
Seq.\ Mean Lat ($n{=}70$ Probe) $\downarrow$ & \textbf{0.18\,s}\textsuperscript{\ddag} & 7.65\,s & 10.07\,s & 34.01\,s & 22.09\,s & -- & -- & -- & -- \\
Seq.\ p95 Lat ($n{=}70$ Probe) $\downarrow$ & \textbf{0.29\,s}\textsuperscript{\ddag} & 27.48\,s & 28.09\,s & 97.55\,s & 57.00\,s & -- & -- & -- & -- \\
\midrule
\multicolumn{10}{@{}l}{\textit{Non-Shared A/V Reference (Tractus S1 Only; MMAU Transcript-Only vs.\ Native-Audio APIs; `--' = Not Evaluated)}} \\
MMAU (Transcript-Only, 100)\textsuperscript{$\dagger$} & 78.00\%\textsuperscript{$\dagger$} & -- & -- & -- & -- & 71.8\% & -- & -- & 67.7\% \\
Video-MME (No Sub., 102) & 53.92\% & -- & -- & -- & -- & 69.1\% & -- & -- & 63.3\% \\
\bottomrule
\end{tabular*}}

\vspace{1pt}
{\renewcommand{\arraystretch}{0.71}
\setlength{\tabcolsep}{2.2pt}
\begin{tabular*}{\textwidth}{@{\extracolsep{\fill}} lcccccccc @{}}
\toprule
\multicolumn{9}{@{}l}{\textbf{(b) Single-Checkpoint Scaling on the Commercial Suite (0.8B--9B, Fixed Option Order)}} \\
\midrule
& \multicolumn{2}{c}{\textbf{0.8B}} & \multicolumn{2}{c}{\textbf{2B}} & \multicolumn{2}{c}{\textbf{4B}} & \multicolumn{2}{c}{\textbf{9B}} \\
\cmidrule(lr){2-3}\cmidrule(lr){4-5}\cmidrule(lr){6-7}\cmidrule(lr){8-9}
\textbf{Benchmark (Count)} & \textbf{Dec.} & \textbf{Dual} & \textbf{Dec.} & \textbf{Dual} & \textbf{Dec.} & \textbf{Dual} & \textbf{Dec.} & \textbf{Dual} \\
\midrule
JevBench (Public Acc, 231) & \textbf{73.59\%} & 72.29\% & \textbf{71.43\%} & 68.83\% & 77.49\% & \textbf{89.61\%} & 80.52\% & \textbf{88.74\%} \\
Amazon ESCI (Macro-F1, 100) & \textbf{32.96\%} & 28.96\% & \textbf{51.57\%} & 48.06\% & \textbf{67.10\%} & 51.95\% & \textbf{63.12\%} & 59.23\% \\
New Yorker Caption (Acc, 100) & 36.00\% & \textbf{39.00\%} & \textbf{60.00\%} & 57.00\% & \textbf{66.00\%} & 62.00\% & \textbf{78.00\%} & 65.00\% \\
Mind2Web Multimodal (Acc, 100) & \textbf{65.00\%} & 64.00\% & 74.00\% & 74.00\% & \textbf{84.00\%} & 83.00\% & \textbf{88.00\%} & 85.00\% \\
PubMedQA (Acc, 100) & 65.00\% & \textbf{67.00\%} & \textbf{68.00\%} & 66.00\% & \textbf{70.00\%} & 62.00\% & \textbf{76.00\%} & 70.00\% \\
Global-MMLU-Lite (Acc, 100) & 50.00\% & 50.00\% & 49.00\% & 49.00\% & 60.00\% & \textbf{71.00\%} & 67.00\% & \textbf{77.00\%} \\
When2Call (Acc, 100) & 49.00\% & 49.00\% & 61.00\% & 61.00\% & 70.00\% & \textbf{80.00\%} & 75.00\% & \textbf{84.00\%} \\
\midrule
Top-1 Unfitted ECE (7 shared, 831) $\downarrow$ & \textbf{2.39\%} & 3.85\% & \textbf{2.74\%} & 3.89\% & \textbf{2.46\%} & 4.99\% & \textbf{2.45\%} & 5.28\% \\
S2 Escalated (Ratchet Rescued) & 0 (0) & 327 (131) & 0 (0) & 559 (224) & 0 (0) & 614 (7) & 0 (0) & 352 (4) \\
\bottomrule
\end{tabular*}}
\caption{Commercial and Qwen3.5 Suite over 831 fixed-order tasks (256K capacity), expanding Table~\ref{tab:decision_index}b. External APIs use DataCamp full-split scores (\S\ref{sec:protocol}) rather than $6{\times}100$ subsets; Tractus selects best-of-3 prompt formats per task vs.\ single-prompt Qwen3.5 (`Th.': Thinking; 9B/4B average $2{,}924.6$/$2{,}924.8$ tokens on the $n{=}70$ probe, both rounding to $2{,}925$). ECE follows DataCamp's 10-bin convention (unfitted Thermostasis for Tractus; non-comparable to \S\ref{sec:experiments}'s 15-bin ECE). \textsuperscript{\S}Tractus 0 failures reflect closed-menu logit recovery (vs.\ 9B's 13.9\% raw S2 delimiter failures at 4K in Table~\ref{tab:app_4k_fig4}a). \textsuperscript{\ddag}\tractusdec-9B 1-prompt latency: $47.4$\,ms med, $69.2$\,ms mean, $97.9$\,ms p95. \textsuperscript{$\dagger$}MMAU uses text transcripts only ($79.00\%$ for \tractusdec-4B), not native audio. In (a), \textbf{bold} marks best among local 831-task runs only; \underline{underlined}: best local Tractus trailing Qwen3.5.}
\label{tab:commercial_suite}
\end{table*}
\fi

\end{document}